\documentclass[runningheads]{llncs}
\usepackage{graphicx}

\usepackage{amsmath,amsfonts,bm}

\def\eqref#1{equation~\ref{#1}}

\def\1{\bm{1}}

\def\vb{{\bm{b}}}
\def\vc{{\bm{c}}}

\def\vf{{\bm{f}}}

\def\vh{{\bm{h}}}
\def\vi{{\bm{i}}}

\def\vo{{\bm{o}}}

\def\vs{{\bm{s}}}

\def\vx{{\bm{x}}}

\def\vz{{\bm{z}}}

\def\evs{{s}}

\def\mM{{\bm{M}}}

\def\mS{{\bm{S}}}

\def\mW{{\bm{W}}}
\def\mX{{\bm{X}}}
\def\mY{{\bm{Y}}}

\DeclareMathAlphabet{\mathsfit}{\encodingdefault}{\sfdefault}{m}{sl}
\SetMathAlphabet{\mathsfit}{bold}{\encodingdefault}{\sfdefault}{bx}{n}

\def\gL{{\mathcal{L}}}
\def\gM{{\mathcal{M}}}

\def\sR{{\mathbb{R}}}

\newcommand{\lr}{\alpha}

\newcommand{\mbeq}{\overset{!}{=}}

\usepackage{amssymb}
\usepackage{amsmath}
\usepackage{amsfonts}
\usepackage{bm}
\usepackage{mathtools}
\usepackage{xspace} 
\usepackage{multirow}
\usepackage{xcolor}
\usepackage{wrapfig}

\usepackage{subcaption}
\usepackage{caption}

\usepackage{pgfplots}
\usepackage{pgfplotstable}
\pgfplotsset{compat=1.13}
\usepgfplotslibrary{fillbetween}
\usepgfplotslibrary{colorbrewer}

\usepackage{anyfontsize}

\usepackage[utf8]{inputenc} 
\usepackage[T1]{fontenc}    
\usepackage{lmodern}

\usepackage{hyperref}       
\usepackage{url}            
\usepackage{booktabs}       
\usepackage{amsfonts}       
\usepackage{nicefrac}       
\usepackage{microtype}      

\usepackage{algorithm}
\usepackage{algorithmic}

\definecolor{darkred}{rgb}{0.5,0,0}
\definecolor{darkgreen}{rgb}{0,0.5,0}
\definecolor{darkblue}{rgb}{0,0,0.5}
\definecolor{lightgrey}{rgb}{0.7,0.7,0.7}
\definecolor{lightergrey}{rgb}{0.93,0.93,0.93}
\definecolor{NavyBlue}{rgb}{0.0, 0.0, 0.5}
\definecolor{OliveGreen}{rgb}{0.33, 0.42, 0.18}

\usepackage{titlecaps}
\Resetlcwords
\Addlcwords{a after along an and are around as at but by etc}
\Addlcwords{for from if in is nor of on the so to with without yet}
\Addlcwords{}

\newcommand{\eg}{e.g., }

\usepackage[toc,title,page]{appendix}

\ifdefined\isoverfull
\else
\fi

\usepackage{tikz}
\usetikzlibrary{calc,decorations.pathmorphing,shapes}
\usetikzlibrary{positioning,shapes,fit,arrows}
\usetikzlibrary{decorations.markings}
\usetikzlibrary{shapes,shapes.geometric}
\usetikzlibrary{shadows,patterns}
\usetikzlibrary{backgrounds,decorations.pathreplacing,automata}
\usetikzlibrary{pgfplots.statistics}

\usepackage{tikz-3dplot}

\usepackage[capitalize]{cleveref}

\newcommand{\tool}{\textsc{Prover}\xspace}

\newcommand{\deeppoly}{\text{DeepPoly}\xspace}

\usepackage[firstpage]{draftwatermark}

\SetWatermarkText{\hspace*{5in}\raisebox{7.5in}{\includegraphics[scale=0.8]{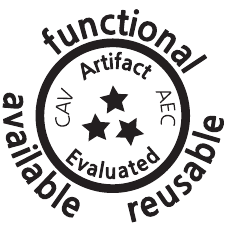}}}

\SetWatermarkAngle{0}

\begin{document}

\title{Scalable Polyhedral Verification of \\ Recurrent Neural Networks}
\author{
    Wonryong Ryou\inst{1} \and
    Jiayu Chen\inst{1} \and
    Mislav Balunovic\inst{1} \and
    Gagandeep Singh\inst{2} \and \\
    Andrei Dan\inst{3} \and
    Martin Vechev\inst{1}
}
\authorrunning{W. Ryou et. al.}
\institute{
    Department of Computer Science, ETH Z\"urich, Switzerland \and
    VMWare Research \&  Department of Computer Science, UIUC, USA \and
    Hitachi Power Grids Research, Switzerland
}
\maketitle

\begin{abstract}
    We present a scalable and precise verifier for recurrent neural networks, called \tool based on two novel ideas: (i)  a method to compute a set of polyhedral abstractions for the non-convex and non-linear recurrent update functions by combining sampling, optimization, and Fermat's theorem, and (ii) a gradient descent based algorithm for abstraction refinement guided by the certification problem that combines multiple abstractions for each neuron. Using \tool, we present the first study of certifying a non-trivial use case of recurrent neural networks, namely speech classification. To achieve this, we additionally develop custom abstractions for the non-linear speech preprocessing pipeline. Our evaluation shows that \tool successfully verifies several challenging recurrent models in computer vision, speech, and motion sensor data classification beyond the reach of prior work.

\end{abstract}

\section{Introduction}

Recurrent neural networks (RNNs) are widely used to model long-term dependencies in lengthy sequential signals 
\cite{diro2018leveraging,openai2018dota,vinyals2019alphastar}. Prior work has demonstrated the susceptibility of RNNs to adversarial perturbations of its inputs \cite{papernot2016crafting}, exposing security vulnerabilities of state-of-the-art RNNs when used in domains such as speech recognition \cite{carlini2018audio,li2019adversarial}, malware detection \cite{hu2017black}, and others. Thus, verifying the robustness of recurrent architectures is critical for their safe deployment. While there has been considerable interest in certifying the robustness of feedforward image classifiers 
\cite{balunovic2019deepg,dvijotham2018training,fischer2020certification,lyu2019fastened,aditi:18,singh2019krelu,singh2019abstract,zhang2018crown}, less attention has been given to recurrent architectures. As a result, current certification solutions do not scale beyond simple models and datasets, which limits their practical applicability. Further, there has been no work on verifying real-world use cases of RNNs. In this paper, we address both of these challenges and present the first precise and scalable verifier for RNNs based on abstract interpretation \cite{cousot1977abstract}, which enables us to certify robustness of realistic speech recognition systems.

\begin{figure}[t]
  \hspace*{-1.8cm}
  \centering
  \scalebox{0.9}{\input{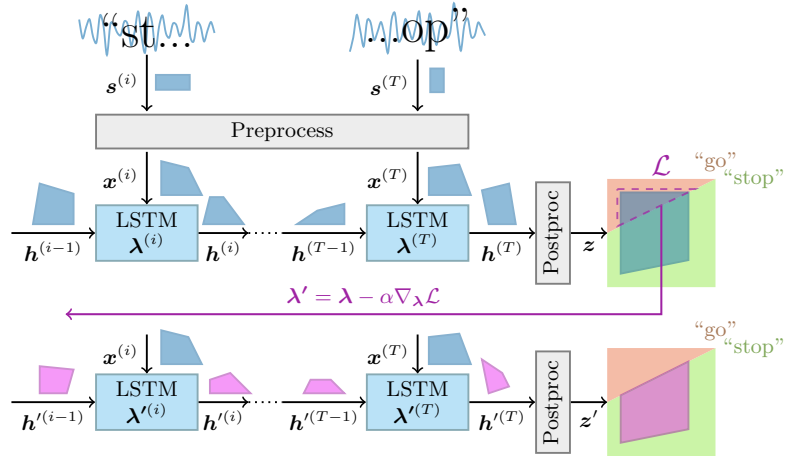}}
  \caption{Certification of recurrent architectures using \tool: utterance ``stop'' with perturbations is correctly classified. Possible perturbations are captured and propagated through the system, then refined backward for improved precision.}
  \label{fig:intro}
  \vspace*{-0.4cm}
\end{figure}

We illustrate the problem setting and overall flow in \cref{fig:intro}. Here, a speech recognition model based on the Long Short-Term Memory (LSTM) architecture \cite{hochreiter1997lstm} receives a signal encoding the utterance of ``stop'' by a human. As such models are usually employed in noisy environments, they must robustly classify variations (e.g., voice changes) to the utterance ``stop''. However, recent work \cite{carlini2018audio} has shown the model may be fooled into classifying the utterance as ``go''. It is important to prove such mis-classifications are not possible, thus avoiding a potential exploitation by an adversary, for instance in automated traffic control settings (which can lead to accidents). Our goal is to design a verifier that can formally establish the robustness of such models against noise-induced perturbations. We focus on LSTMs, as they are the most widely used form of RNNs, but our methodology can be easily extended to other architectures (e.g., Gated Recurrent Unit (GRU) \cite{DBLP:conf/emnlp/ChoMGBBSB14}). \cref{fig:intro} shows how our proposed verifier, called \tool, (Polyhedral Robustness Verifier of RNNs) automatically verifies the robustness of the model. Here, the labeled rectangles represent operations in the network. The ``Preprocess'' box captures domain-specific pre-processing operations (typically present when using RNNs, e.g., speech processing). In our method, we first compute a polyhedral abstraction capturing all speech signals given as input to the model under the given perturbation budget. At each timestep $i$, the pre-processing operation receives a polyhedron $\vs^{(i)}$ and produces an output polyhedron $\vx^{(i)}$. This shape is then propagated symbolically through the LSTM and the post-processing stage, resulting in a polyhedral output shape, denoted as $\vz$ (blue shape in~\cref{fig:intro}).

\textbf{Key challenge: polyhedral abstractions for LSTMs}
The main challenge in certifying LSTMs is the design of precise and scalable polyhedral abstract transformers for the non-linear operations employed in LSTMs: given a polyhedral shape capturing hidden states $\vh^{(i-1)}$, to produce the shape capturing the next set of hidden states $\vh^{(i)}$. A recent method \cite{ko2019popqorn} computes this based on gradient-based optimization but suffers from two main limitations. First, the optimization procedure is computationally expensive and does not scale to realistic use cases. Second, the method lacks convergence and optimality guarantees. To address these issues, we introduce a novel technique based on a combination of sampling, linear programming, and Fermat's theorem  \cite{fermat_theorem}, which significantly improves the precision and scalability compared to prior work \cite{ko2019popqorn}, while offering asymptotic guarantees of convergence towards the optimal solution.

\textbf{Refinement via optimization}
To certify robustness, we must verify that each concrete point in the output shape $\vz$ corresponds to the correct label ``stop''. However, $\vz$ can contain, due to over-approximation, spurious incorrect concrete points (it intersects the red region representing incorrect outputs). To address this issue, we form a loss based on the output shape, backpropagate the gradient of this loss through the timesteps and adjust the polyhedral abstractions in each LSTM unit to decrease the loss. The goal is to refine the abstraction, guided by the certification task. We illustrate this process in ~\cref{fig:intro} using the purple backward arrow with the refined polyhedral abstraction shown in purple.
Using the refined abstraction, the new output shape $\vz'$ (purple polygon) lies completely inside the green region of the output space, meaning it provably contains only correct output vectors (corresponding to ``stop''), and hence certification succeeds. Overall, our method significantly increases the precision of end-to-end RNN certification without introducing high runtime costs.

\textbf{Key contributions} Our main contributions are:
\begin{itemize}
  \item A new and efficient method to certify the robustness of RNNs to adversarial perturbations. Our method relies on novel polyhedral abstractions for handling non-linear operations in these architectures.
  \item A novel method that automatically refines the abstraction for each input example being certified guided by the certification task.
  \item An implementation of the method in a system called \tool and evaluation on several benchmarks and datasets. Our results show that \tool is precise and scales to larger models than prior work. \tool is also the first verifier able to certify realistic RNN-based speech classifiers. The code is available in \url{https://github.com/eth-sri/prover}.
\end{itemize}

\section{Related work}

While the first adversarial examples for neural networks were found in computer vision \cite{szegedy2013intriguing,biggio2013evasion}, recent work also showed the vulnerability of RNNs \cite{papernot2016crafting}. Modern speech recognition systems, based on RNNs, were shown susceptible to small noise crafted by an adversary using white-box attacks \cite{carlini2016hidden,carlini2018audio}, achieving a 100\% success rate against DeepSpeech \cite{hannun2014deep}, a state-of-the-art speech-to-text engine. These were later followed by attacks based on universal perturbation \cite{neekhara2019universal} and temporal dependency \cite{yang2019characterizing}. Recent work \cite{qin2019imperceptible,li2019adversarial} demonstrates that adversarial examples for audio classifiers are realizable in the real-world. While giving an empirical estimate of the vulnerability of RNNs, these works do not provide any formal guarantees, which is the goal of our work.


There have also been recent works on the verification of RNNs. \cite{akintunde2019verification} propose the certification of RNNs based on mixed-integer linear programming, which only works for ReLU-based networks and does not consider LSTMs, which use sigmoid and tanh activations. \cite{wu2019robustness} propose an input discretization method to certify video models that are a combination of CNNs and RNNs. However, discretization does not scale to the perturbations we consider in our work. \cite{jacoby2020verifying} propose to verify RNNs by automatically inferring temporal homogeneous invariants using binary search. However, their approach is limited to vanilla RNNs and does not apply to the more commonly used LSTM networks considered in this work. \cite{khmelnitsky2020property} propose the statistical variant of Angulin's algorithm \cite{angluin1987learning} for probabilistic verification and counterexample generation for RNNs, however they cannot provide deterministic guarantees as our work. The work most related to ours is POPQORN \cite{ko2019popqorn} which uses expensive gradient-based optimizations for every operation in the network. We experimentally show that it does not scale to practical applications such as speech classification.

\section{Background} \label{sec:background}

We first define the threat model and then present all operations that are part of the verification procedure, including speech preprocessing and LSTM updates.

\subsection{Threat model} \label{sec:threat_model}
We use a threat model based on the $L_\infty$-norm, where an attacker can change each element of a correctly classified input vector $\vs$ by an amount $\leq \epsilon \in \mathbb{R}$ \cite{carlini2018audio}.
Therefore, our input region can be represented as a conjunction of intervals $[s_i -\epsilon, s_i+\epsilon]$, where $s_i$ is the $i$-th element of $\vs$.
The measure of signal distortion in this setting are decibels (dB) defined as:
\[ dB(\vs) = \max_i 20 \cdot \log_{10} (|\evs_i|);\ dB_s(\bm\delta) = dB(\bm\delta) - dB(\vs) \]
The quieter the perturbation is, the smaller $dB_s(\bm\delta)$ is.
We fix the $dB_s(\bm\delta) =: \epsilon$ as \textit{dB perturbation} and focus on
verifying that the model classifies correctly all signals $\vs'$ possible under our threat model.

\begin{figure}[t]
  \centering
  \scalebox{0.85}{
\usetikzlibrary{positioning, fit, arrows.meta, shapes}

\newcommand{\empt}[2]{$#1^{(#2)}$}

\begin{tikzpicture}[
    font=\sf \scriptsize,
    >=LaTeX,
    cell/.style={
        rectangle,
        rounded corners=5mm,
        draw,
        very thick,
        },
    operator/.style={
        circle,
        draw,
        inner sep=-0.5pt,
        minimum height =.4cm,
        },
    function/.style={
        ellipse,
        draw,
        inner sep=1pt
        },
    ct/.style={
        line width = .75pt,
        minimum width=1cm,
        minimum height=0.6cm,
        inner sep=1pt,
        },
    gt/.style={
        rectangle,
        draw,
        minimum width=4mm,
        minimum height=3mm,
        inner sep=1pt
        },
    mylabel/.style={
        font=\normalsize
        },
    ArrowC1/.style={
        rounded corners=.25cm,
        thick,
        },
    ArrowC2/.style={
        rounded corners=.5cm,
        thick,
        },
    ]

    \node [cell, minimum height=3.5cm, minimum width=6cm] at (0,0){} ;

    \node [gt] (ibox1) at (-2,-0.75) {$\sigma$};
    \node [gt] (ibox2) at (-1.5,-0.75) {$\sigma$};
    \node [gt, minimum width=1cm] (ibox3) at (-0.5,-0.75) {Tanh};
    \node [gt] (ibox4) at (0.5,-0.75) {$\sigma$};

    \node [operator] (mux1) at (-2,1.5) {$\times$};
    \node [operator] (add1) at (-0.5,1.5) {+};
    \node [operator] (mux2) at (-0.5,0) {$\times$};
    \node [operator] (mux3) at (1.5,0) {$\times$};
    \node [gt, minimum width=1cm] (func1) at (1.5,0.75) {Tanh};

    \node[ct, label={[mylabel]left:Prev. cell state}] (c) at (-4,1.5) {\empt{\vc}{t-1}};
    \node[ct, label={[mylabel]left:Prev. hidden state}] (h) at (-4,-1.5) {\empt{\vh}{t-1}};
    \node[ct, label={[mylabel]left:Input}] (x) at (-2.5,-2.3) {\empt{\vx}{t}};

    \node[ct] (f) at (-2.3,-1.2) {\empt{\vf_0}{t}};
    \node[ct] (i) at (-1.7,-1.2) {\empt{\vi_0}{t}};
    \node[ct] (cc) at (-0.7,-1.2) {\empt{\tilde{\vc}_0}{t}};
    \node[ct] (o) at (0.2,-1.2) {\empt{\tilde{\vo}_0}{t}};

    \node[ct, label={[mylabel]right:Cell state}] (c2) at (4,1.5) {\empt{\vc}{t}};
    \node[ct, label={[mylabel]right:Hidden state}] (h2) at (4,-1.5) {\empt{\vh}{t}};
    \node[ct, label={[mylabel]right:Cell output}] (x2) at (2.5,2.3) {\empt{\vh}{t}};

    \draw [ArrowC1] (c) -- (mux1) -- (add1) -- (c2);

    \draw [ArrowC2] (h) -| (ibox4);
    \draw [ArrowC1] (h -| ibox1)++(-0.5,0) -| (ibox1);
    \draw [ArrowC1] (h -| ibox2)++(-0.5,0) -| (ibox2);
    \draw [ArrowC1] (h -| ibox3)++(-0.5,0) -| (ibox3);
    \draw [ArrowC1] (x) -- (x |- h)-| (ibox3);

    \draw [->, ArrowC2] (ibox1) -- (mux1);
    \draw [->, ArrowC2] (ibox2) |- (mux2);
    \draw [->, ArrowC2] (ibox3) -- (mux2);
    \draw [->, ArrowC2] (ibox4) |- (mux3);
    \draw [->, ArrowC2] (mux2) -- (add1);
    \draw [-, ArrowC1] (add1 -| func1)++(-0.5,0) -| (func1);
    \draw [->, ArrowC2] (func1) -- (mux3);

    \draw [-, ArrowC2] (mux3) |- (h2);
    \draw (c2 -| x2) ++(0,-0.1) coordinate (i1);
    \draw [-, ArrowC2] (h2 -| x2)++(-0.5,0) -| (i1);
    \draw [-, ArrowC2] (i1)++(0,0.2) -- (x2);

\end{tikzpicture}}
  \caption{LSTM cell: $\vf_0^{(t)}, \vi_0^{(t)}, \vo_0^{(t)}$, and $\tilde{\vc}_0^{(t)}$ represent the pre-activated gates.}
  \label{fig:lstm_cell}
\end{figure}
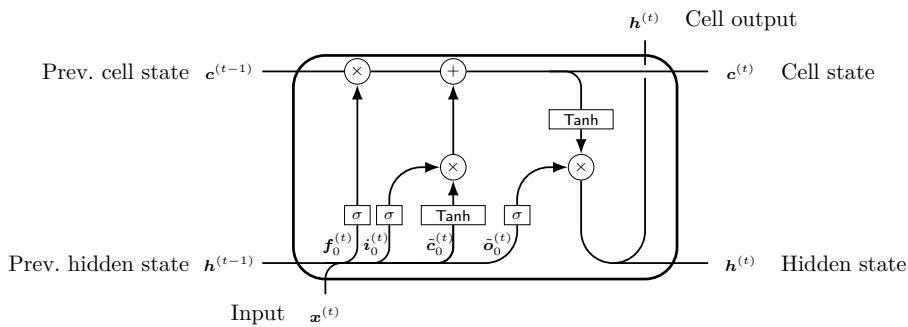

\subsection{Long Short-Term Memory (LSTM)} \label{sec:lstm_explained}

LSTM architectures \cite{hochreiter1997lstm} are popular for handling sequential data as they can utilize long-term dependencies. These dependencies are passed through time using two state vectors for the timestep $t$: cell state $\vc^{(t)}$ and hidden state $\vh^{(t)}$. These state vectors are updated using the following formulas:
\begin{equation*}
  \begin{aligned}[c]
    \vf_0^{(t)} &= [\vx^{(t)}, \vh^{(t-1)}] \mW_f + \vb_f \\
    \vo_0^{(t)} &= [\vx^{(t)}, \vh^{(t-1)}] \mW_o + \vb_o \\
    \vc^{(t)} &= \sigma (\vf_0^{(t)}) \odot \vc^{(t-1)} + \sigma(\vi_0^{(t)}) \odot \tanh (\tilde{\vc}_0^{(t)}) \\
  \end{aligned}
  \qquad
  \begin{aligned}[c]
    \vi_0^{(t)} &= [\vx^{(t)}, \vh^{(t-1)}] \mW_i + \vb_i \\
    \tilde{\vc}_0^{(t)} &= [\vx^{(t)}, \vh^{(t-1)}] \mW_{\tilde{c}} + \vb_{\tilde{c}} \\
    \vh^{(t)} &= \sigma (\vo_0^{(t)}) \odot \tanh (\vc^{(t)})
  \end{aligned}
\end{equation*}
where $[\cdot, \cdot]$ is the horizontal concatenation of two row vectors, $\mW_\cdot$ and $\vb_\cdot$ are the kernel and bias of the cell, respectively, and $\sigma$ is the sigmoid function. At timestep $t$, vectors $\vf_0^{(t)}, \vi_0^{(t)}, \vo_0^{(t)}, \tilde{\vc}_0^{(t)}$ represent pre-activations of the forget gate, input gate, output gate and the candidate gate, respectively. We show an illustration of an LSTM cell in \cref{fig:lstm_cell}. We treat $\sigma$ and $\tanh$ as forms of activation functions, which is why we define the LSTM using pre-activations.

Intuitively, the \textit{input gate} transforms the input vector, the \textit{forget gate} filters the information from the previous cell state, the \textit{candidate gate} prepares the candidate cell state, and the \textit{output gate} transforms the current hidden state. All of these gates receive as input the hidden state $\vh^{(t-1)}$ of the previous cell and the input $\vx^{(t)}$ representing the current frame. This recurrent architecture allows inputs with arbitrary length, enabling LSTMs to handle temporal data, e.g., speech processing.

\subsection{Speech preprocessing} \label{sec:mel_explained}

Though there have been various works that operate directly on the raw signal \cite{pascual2017segan,sainath2015learning}, speech signals are commonly preprocessed using the \emph{filterbank} or \textit{log Mel-filterbank energy} methods. The result is a vector of coefficients whose elements contain log-scaled values of filtered spectra, one for every Mel-frequency. This method models the non-linear human acoustic perception as power spectrum filters based on Mel-frequencies. The input signal is split into several (possibly overlapping) frames for granular analysis, and the following steps are applied:
\begin{enumerate}
  \item \textit{Pre-emphasizing and windowing} are preprocessing stages on the raw signals. Speech signals tend to have larger and smoother low-frequency samples and smaller and fluctuating high-frequency samples. \textit{Pre-emphasizing} is a process of subtracting the adjacent sampled values multiplied by a scalar parameter ($\vs^{(i)}_j - \alpha \vs^{(i)}_{j-1}$, commonly \(\alpha=0.97\)). This alleviates the unbalanced distribution of signal strength along with the frequency. \textit{Windowing} involves multiplication of each sampled value and `windows' according to their indices. The window here refers to a Hamming window, which is a bell-like curve with peak in the middle of the frame and drops at the side. It reduces the border effects on each frame by suppressing the values near the border with smaller values.
  
  \item \textit{Power spectrum of Fast Fourier transform (FFT)} performs the discrete Fourier transform (DFT) and obtains the squared norm of each element to obtain intensities in the frequency domain. FFT consists of matrix multiplications with complex entries. We modify it to use only real numbers by: (i) separating real and imaginary parts of the matrix and constructing two separate matrices, (ii) multiplying each matrix with the signal, (iii) squaring the entries, and (iv) adding the resulting matrices entry-wise.
  
  \item \textit{Mel-filter bank log energy:} The Mel-frequency filters are triangular, each emphasizing the power of the selected frequency and suppressing the adjacent ones. In our case, we (i) apply the Mel-filterbank to the power spectrum and (ii) take the log of the entries to adjust the level.
\end{enumerate}
Following \cite{sahidullah2012design}, each step can be represented as a distinct matrix operation. It allows us to decompose and rearrange the steps into slightly different stages:
\begin{enumerate}
  \item \textit{Pre-square stage: $\mS \rightarrow \mY = \mS \mM_{1}$}. This stage contains pre-emphasizing, windowing (step 1), and FFT (until step 2-(ii)). All operations are representable as matrix multiplications, so we pre-calculate the product matrix.
  \item \textit{Square stage: $\mY \rightarrow \bm{\theta} = \mY \odot \mY$}. This is step 2-(iii). Entry-wise square operations cannot be combined with other matrix multiplications.
  \item \textit{Pre-log stage: $\bm{\theta} \rightarrow \tilde{\mX} = \bm{\theta} \mM_{2}$}. From step 2-(iv) through step 3-(i). We combine the operations into a single matrix.
  \item \textit{Log stage: $\tilde{\mX} \rightarrow \mX = \log \tilde{\mX}$}. Applying entry-wise logarithm (step 3-(ii)).
\end{enumerate}
We use the resulting $\mX = {[\vx^{(1)}\ \cdots\ \vx^{(T)}]}^\intercal$ as the input to the neural network.

\subsection{Verification using \deeppoly abstract domain} \label{sec:deeppoly}

\deeppoly \cite{singh2019abstract} is a sub-polyhedral abstract domain that associates a lower and an upper polyhedral bound and interval bounds per neuron. It is faster than Polyhedra \cite{Singh:17} and more precise than other weakly relational domains such as Octagons \cite{Mine:06}, Zones \cite{Mine:01}, and Zonotopes \cite{singh2018deepz} when analyzing neural networks. Previously, it has been suceesfully applied for verifying feedforward networks in \cite{singh2019abstract,balunovic2019deepg}. Formally, let $\mathcal{X}=\{x_1,x_2,\hdots,x_n\}$ be an ordered set of neurons such that the neurons in layer $l$ appear before the neurons of layer $l'>l$. \deeppoly associates with each neuron $x_j$, both interval $l_j \leq x_j \leq u_j$ and polyhedral bounds $\sum_{i<j} a_i \cdot x_i + b \leq x_j \leq \sum_{i<j} a'_i \cdot x_i + b'$ where $l_j, u_j, a_i, a'_i, b,b' \in \mathbb{R} \cup \{\infty\}$.  \deeppoly is exact for affine transformations which are frequently applied both in the speech preprocessing pipeline and the LSTM unit. \deeppoly loses precision for the non-linear operations in LSTMs. We note that computing polyhedral bounds on their output is more challenging than for feedforward networks.

The precision of the \deeppoly approximation for the non-linear operations depends on the tightness of the interval bounds of the neurons that are input to the non-linear operations. \deeppoly provides a scalable and precise method called \textit{backsubstitution} for optimizing a linear expression within a region defined by the set of \deeppoly constraints. It does so by recursively substituting the bounding linear expressions of target neurons with the polyhedral bounds of previous layers' neurons until reaching the input neurons. It then uses the concrete bounds of the input neurons for computing the result. Backsubstitution is used for computing the interval bounds of neurons input to the non-linear operations as well as for bounding the difference between the neurons in the output layer needed to prove robustness.
We refer the reader to \cite{singh2019abstract} for details of the backsubstitution.


\section{Overview of \tool} \label{sec:overview}

This section illustrates the workings of \tool on a small example. Our goal is to certify the robustness of a single LSTM cell on the input $x \in [-1.2, 1.2]$. For this example, we assume that there are two output classes and all intermediate LSTM gates $\{\vi,\vf,\tilde\vc,\vo\}$ share the same weights and biases:
\begin{align*}
  \{\vi,\vf,\tilde\vc,\vo\} &= \left[\begin{matrix} 1 \\ 0.5 \end{matrix} \right] x + \left[\begin{matrix} 0 \\ 1 \end{matrix} \right],
  \qquad \vc = \sigma(\vi) \odot \tanh(\tilde\vc),
  \qquad \vh  = \sigma(\vo) \odot \tanh(\vc).
\end{align*}

The correct output here is $h_2$ and to certify robustness we need to prove that $h_2 - h_1 > 0$ holds for all inputs $x$. In other words, $\min h_2 - h_1 > 0$.


\textbf{Polyhedral abstraction}
We build our verifier based on the \deeppoly \cite{singh2019abstract} abstraction since \deeppoly outperforms the interval analysis and other competitive domains, as \cref{sec:deeppoly} states.

\textbf{Challenges in computing polyhedral bounds for LSTMs}
The composed binary non-linear operations applied in LSTMs such as $\sigma(x)\tanh(y)$ and $\sigma(x)y$ are significantly more complex to handle than the ReLU, Sigmoid, and Tanh activations originally handled by \cite{singh2019abstract}. This is because the non-linear operations in LSTMs mentioned above involve transcendental functions yielding non-linear 3D curves that are neither convex nor concave. The optimal polyhedral bounds for these operations have no closed-form solution and cannot be calculated by simple geometry or algebra. Further, obtaining such bounds is computationally expensive \cite{ko2019popqorn}. For example, obtaining the lower linear plane for bounding $\sigma(x)\tanh(y)$ is equivalent to solving a Lagrangian with 6 variables - 3 linear coefficients, 2 interval-bounded coordinates and 1 Lagrange multiplier for the constraint. In contrast, the optimal polyhedral bounds for ReLU, Sigmoid, and Tanh have closed form solutions, easily visualized in 2D.

\textbf{Precise polyhedral bounds via LP} 
To overcome these challenges, we propose a generic approach based on linear programming (LP) to compute precise polyhedral bounds. We illustrate our approach for calculating a lower polyhedral bound of $h_2 = \sigma(o_2)\tanh(c_2)$. First, we calculate the concrete intervals for the two target variables via \textit{backsubstitution} \cite{singh2019abstract}, briefly described in \cref{sec:deeppoly}.
In our case, the target variables are $o_2$ and $c_2$ and the backsubstitution yields $o_2 \in [0.4, 1.6]$ and $c_2 \in [-0.79, 0.62]$. Our abstraction can represent the affine transformations exactly. Therefore, we obtain the exact interval for $o_2 = 0.5 \cdot x+1$ via the backsubstitution whereas the obtained interval for $c_2$ is an overapproximate one. Then, we uniformly sample a set of points $\{(x_1, y_1), ..., (x_n, y_n)\}$ from the input domain $[0.4, 1.6] \times [-0.79, 0.62]$. We solve the following optimization problem to calculate the lower polyhedral bound of $h_2$:
\begin{equation*}
  \min_{A_l, B_l, C_l \in \sR} \sum_{i=1}^n \left( \sigma(x_i)\tanh(y_i) - (A_l \cdot x_i + B_l \cdot y_i + C_l) \right),
\end{equation*}
subject to the constraint that $A_l \cdot x_i + B_l \cdot y_i + C_l \leq \sigma(x_i)\tanh(y_i)$ for each $i$. This is a linear program over three variables ($A_l, B_l, C_l$) that can be solved efficiently in polynomial time. However, the obtained bound may not be sound as the sampled points do not fully cover the continuous input domain. To address this, we shift the plane downwards by an offset (decreasing $C_l$) equal to the maximum violation between $A_l \cdot x + B_l \cdot y + C_l$ and $h_2$ based on Fermat's theorem. After solving the linear program and the adjustment, we obtain $A_l = 0.04, B_l = 0.46, C_l = 0.01$ which results in the following lower polyhedral bound to $h_2$: $h_2 \geq LB_{h_2} = 0.04 \cdot o_2 + 0.46 \cdot c_2 + 0.01$. We compute the upper bound to $h_2:  h_2 \leq UB_{h_2}$ analogously. After computing a polyhedral abstraction of each neuron, we calculate the lower bound of $h_2 - h_1$ via backsubstitution as follows:
\begin{align*}
  &\min h_2 - h_1 \geq LB_{h_2} - UB_{h_1} \\
  &\geq (0.04 \cdot o_2 + 0.46 \cdot c_2 + 0.01) - (-0.09 \cdot o_1 + 0.66 \cdot c_1 + 0.14) \\
  &\geq 0.04 \cdot o_2 + 0.46 \cdot (0.07 \cdot i_2 + 0.27 \cdot g_2 + 0.09) \\
  &\qquad + 0.09 \cdot o_1 - 0.66 \cdot (-0.04 \cdot i_1 + 0.38 \cdot g_1 + 0.25) - 0.14 \\
  &\geq 0.20 \cdot (0.5 \cdot x+1) -0.13 \cdot x - 0.10
  \geq -0.03 \cdot x -0.08 \geq -0.11.
\end{align*}

The precision of the bounds generated by our LP-based method increases with the number of samples yielding optimal bounds (in the sense of small gap) asymptotically. For our example, the computed bounds are optimal.

While our optimal bounds significantly improve precision compared to intervals, they are not sufficient to certify robustness. Prior work for ReLU networks \cite{dvijotham2018training,balunovic2020colt,lyu2019fastened} showed that the greedy approach of always selecting the optimal bounds minimizing the gap can yield less precise results than an adaptive strategy which computes bounds guided by the certification problem. Based on this observation, we introduce a novel approach based on splitting and gradient descent that computes polyhedral abstractions for non-linearities employed in LSTMs informed by the certification problem and proves that $\min h_2 - h_1 > 0$ actually holds.

\textbf{Abstraction refinement via splitting and gradient descent}
While our method based on LP offers an efficient way to compute polyhedral abstraction of activation functions, its main limitation is that the abstraction cannot be refined based on the certification goal. In this work, we introduce a novel method where we first compute mutiple sound bounds for the neuron using our LP method and then automatically obtain a combination of the computed bounds that improves the lower bound of our certification objective $h_2 - h_1$ for each input example. As before, we use the backsubstitution to obtain the interval bounds for the input variables. Since the output of our LP method is sensitive to the choice of the sampled points, we split the original input region $[l_x, u_x] \times [l_y, u_y]$ to sample more effectively from smaller sub-regions thereby reducing the chances of missing an outlier. We found that splitting along the two diagonals of $[l_x, u_x] \times [l_y, u_y]$ into four triangular zones, denoted as $\mathcal{T}_k$, $k \in \{1, 2, 3, 4\}$, performs the best in our evaluation. We use $\mathcal{T}_0$ to denote the original input region. Next, we calculate four additional planes, for both the upper and lower bounds, by sampling each subregion $\mathcal{T}_k$ and then applying our LP method as before. We refer to each plane as a \textit{candidate bound}:
\begin{align*}
  \min_{A_l, B_l, C_l \in \sR} & \sum_{i=1}^{n} \left( \sigma(x_i)\tanh(y_i) - (A_l \cdot x_i + B_l \cdot y_i + C_l) \right) \\
  \text{subject to } & \bigwedge_{i=1}^n A_l \cdot x_i + B_l \cdot y_i + C_l \leq \sigma(x_i)\tanh(y_i) \hspace{1mm}
  \text{where }  (x_i, y_i) \sim \mathcal{T}_k 
\end{align*}

Using our LP based method, we obtain the following corresponding candidate polyhedral abstraction for $h_2$, $LB^k_{h_2}$ for each $\mathcal{T}_k$ in our example:
\begin{align*}
  h_2 \geq LB^0_{h_2} &= 0.04 \cdot o_1 + 0.46 \cdot c_1 + 0.01,
  h_2 \geq LB^1_{h_2} = 0.04 \cdot o_1 + 0.46 \cdot c_1 + 0.01 \\
  h_2 \geq LB^2_{h_2} &= 0.13 \cdot o_1 + 0.63 \cdot c_1 - 0.17,
  h_2 \geq LB^3_{h_2} = 0.04 \cdot o_1 + 0.46 \cdot c_1 + 0.01 \\
  h_2 \geq LB^4_{h_2} &= 0.13 \cdot o_1 + 0.63 \cdot c_1 - 0.17
\end{align*}

Note that $LB^0_{h_2}$ denotes the polyhedral abstraction calculated for the whole region, and there might be duplicate $LB$'s when the curve in the given subregion is concave. The final bound $LB_{h_2}$ is a linear combination of $LB^k_{h_2}$:
\[ LB_{h_2} = \sum_{k=0}^4 \lambda_i \cdot LB_{h_2}^k, \qquad \sum_{k=0}^4 \lambda_i = 1. \]
Our optimization algorithm, explained in~\cref{sec:opt_bounds}, learns the values of $\lambda_i$ via gradient descent that maximizes $\min h_2 -h_1$. For our example, we obtain $\mathbb{\lambda} = (0.09, 0.13, 0.34, 0.09, 0.35)$ as the set of coefficients which results in a new lower bound of $h_2 \geq 0.10 \cdot o_2 + 0.58 \cdot c_2 - 0.11$ for the neuron $h_2$. We improve the bounds for other neurons in a similar fashion. Using the new bounds, we obtain $h_2 - h_1 \geq 0.01 > 0$ which enables us to correctly certify the predicate of interest. If the certification still fails, it is possible to further refine the abstraction by increasing the number of splits and repeating the procedure above.

Compared to \cite{ko2019popqorn}, which uses a single bound, our method is more flexible and can tune $\bm{\lambda}$ parameters to find a combination of different bounds for each neuron that yields the most precise certification result for each certification instance. Our method is also faster as it performs expensive gradient-based optimization for only the output layer whereas \cite{ko2019popqorn} performs this step for each neuron in the LSTM twice. \cite{dvijotham2018training,lyu2019fastened,balunovic2020colt} also suggest a similar idea of bounding ReLU's lower bound using gradient descent, but their approach is limited to unary functions with trivial candidates, not applicable to our setting which requires handling complex binary operations with non-trivial initial bounds.

\textbf{Generality of our method}
Our method is generic and can be easily extended to obtain polyhedral bounds for the non-linear operations in other architectures such as transformers \cite{vaswani:17} and capsule networks \cite{capsule:17}. 

\section{Scalable Certification of LSTMs} \label{sec:lstmcrt}

Next, we formally describe our scalable verifier for LSTM networks. As mentioned in \cref{sec:overview}, we build our verifier based on the \deeppoly abstract domain \cite{singh2019abstract} introduced in \cref{sec:deeppoly}. For simplicity, we focus on computing the polyhedral bounds for the output of non-linear operations. Note that the computed polyhedral bounds contain only the neurons from the previous layers. This restriction is required for backsubstitution used for computing the interval bounds of the inputs, which is an approximate algorithm for solving an LP (e.g. maximize or minimize $x_j$) within a polyhedral region defined by \deeppoly constraints. In \cref{sec:sampling_lp}, we show how to obtain tight, asymptotically optimal polyhedral bounds on key operations in the LSTM unit: $\sigma(x)\tanh(y)$ and $\sigma(x)y$. \cref{sec:opt_bounds} describes a novel method to dynamically choose between different polyhedral bounds for increasing verifier precision.


\subsection{Computing polyhedral abstractions of LSTM operations} \label{sec:sampling_lp}

Our goal is to bound the products of \textit{sigmoid and tanh} and \textit{sigmoid and identity}, using lower and upper polyhedral planes parameterized by coefficients $A_l$, $B_l$, $C_l$ and $A_u$, $B_u$, $C_u$, respectively. Let $f(x, y) = \sigma(x) \tanh(y)$ and $g(x, y) = \sigma(x) y$. For $h \in \{f, g\}$ we describe how to obtain the lower and upper bounds of $h$:
\begin{equation*}
  A_l \cdot x + B_l \cdot y + C_l \leq h(x, y) \leq A_u \cdot x + B_u \cdot y + C_u
\end{equation*}

We formulate the search for a lower bound of $h(x, y)$ as an optimization problem that minimizes the volume between the bound and the function, subject to the (soundness) constraint that the lower bound is below the function value:
\begin{align} \label{eq:int_formulation}
  \min_{A_l, B_l, C_l} & \int_{(x, y) \in B} \left( h(x, y) - (A_l \cdot x + B_l \cdot y + C_l) \right) \nonumber \\
  \text{subject to } & A_l \cdot x + B_l \cdot y + C_l \leq h(x, y), \forall (x, y) \in B.
\end{align}

We denote $B = [l_x, u_x] \times [l_y, u_y]$ as the boundaries of input neurons $x$ and $y$ obtained using backsubstitution. We next describe our method to solve \cref{eq:int_formulation}.

\textbf{Step 1: Approximation via LP}
We solve an approximation of the intractable optimization problem from \cref{eq:int_formulation}, obtaining potentially unsound constraints. Unsoundness implies that there can be points in region $B$ which violate the bounds. We build on the approach from \cite{balunovic2019deepg}, which proposes to approximate the objective in \cref{eq:int_formulation} using Monte Carlo sampling. Let $D = \{(x_1, y_1), \ldots, (x_n, y_n)\}$ be a set of points from $B$ sampled uniformly at random. We phrase the following optimization problem:
\begin{align} \label{eq:lp_formulation2}
  \min_{A_l, B_l, C_l \in \sR} & \sum_{i=1}^n \left( h(x_i, y_i) - (A_l \cdot x_i + B_l \cdot y_i + C_l) \right) \nonumber \\
  \text{subject to } & \bigwedge_{i=1}^n A_l \cdot x_i + B_l \cdot y_i + C_l \leq h(x_i, y_i).
\end{align}

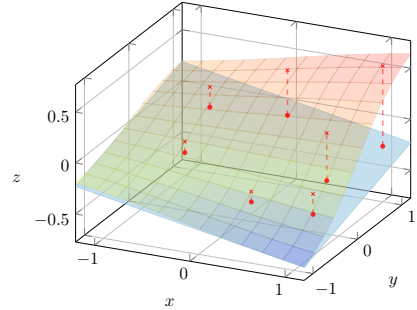
\begin{wrapfigure}{r}{0.45\textwidth}
\vspace{-33pt}
  \centering
  \scalebox{0.65}{\definecolor{lower_col}{RGB}{52,111,72}
\definecolor{upper_col}{RGB}{158,202,225}

\tikzstyle{bnds}=[opacity=0.5, no markers, dashed, line width=0.3mm, draw=red]

\begin{tikzpicture}
  [declare function={
  lb(\x,\y) = -0.17425786836211618*\x+0.1730982178929485*\y-0.2297638790769432;
  curve(\x,\y) = tanh(\y) / (1 + exp(-\x));
  },
  lbsty/.style = {rotate=180}]

  \begin{axis}[domain=-1.2:1.2, domain y=-1.2:1.2, grid=both,
    xlabel=$x$,
    ylabel=$y$,
    zlabel=$z$,
    zlabel style={rotate = -90}
    ]
    \addplot3[surf, samples=2, opacity=0.8, fill opacity=0.6, color=upper_col, faceted color=upper_col]
      {lb(x,y)};

    \addplot3+ [bnds] coordinates {
    (0.7855104790072773, 0.19923964128199212, -0.33215723388531304)
    (0.7855104790072773, 0.19923964128199212, 0.13506850727576195)};

    \addplot3+ [bnds] coordinates {
    (0.018649526493345725, 0.9582560751985689, -0.06714128690767293)
    (0.018649526493345725, 0.9582560751985689, 0.3752153588983146)};

    \addplot3+ [bnds] coordinates {
    (-0.5509358644347242, -0.13160359959580847, -0.15653931829463996)
    (-0.5509358644347242, -0.13160359959580847, -0.04784460539110303)};

    \addplot3+ [bnds] coordinates {
    (0.3247395101781414, -0.5112923109937402, -0.3748560817489275)
    (0.3247395101781414, -0.5112923109937402, -0.2733773910534693)};

    \addplot3+ [bnds] coordinates {
    (0.9071566653657073, -0.3703423474683789, -0.4519486662111244)
    (0.9071566653657073, -0.3703423474683789, -0.25240336562594695)};

    \addplot3+ [bnds] coordinates {
    (-0.6623296448437159, 0.6712949089831381, 0.0018522254121504655)
    (-0.6623296448437159, 0.6712949089831381, 0.19930937188428607)};

    \addplot3+ [bnds] coordinates {
    (1, 1, -0.2309235295461109)
    (1, 1, 0.5567699411459397)};

    \addplot3+ [opacity=0.8, fill=red, draw=red, mark=*, mark size=0.4mm, only marks] coordinates {
    (0.7855104790072773, 0.19923964128199212, -0.33215723388531304)
    (0.018649526493345725, 0.9582560751985689, -0.06714128690767293)
    (-0.5509358644347242, -0.13160359959580847, -0.15653931829463996)
    (0.3247395101781414, -0.5112923109937402, -0.3748560817489275)
    (0.9071566653657073, -0.3703423474683789, -0.4519486662111244)
    (-0.6623296448437159, 0.6712949089831381, 0.0018522254121504655)
    (1, 1, -0.2309235295461109)
    };

    \addplot3+ [opacity=0.8, draw=red, mark=x, mark size=0.5mm, only marks] coordinates {
    (0.7855104790072773, 0.19923964128199212, 0.13506850727576195)
    (0.018649526493345725, 0.9582560751985689, 0.3752153588983146)
    (-0.5509358644347242, -0.13160359959580847, -0.04784460539110303)
    (0.3247395101781414, -0.5112923109937402, -0.2733773910534693)
    (0.9071566653657073, -0.3703423474683789, -0.25240336562594695)
    (-0.6623296448437159, 0.6712949089831381, 0.19930937188428607)
    (1, 1, 0.5567699411459397)
    };

    \addplot3[surf, samples=12, opacity=0.2] {curve(x,y)};

  \end{axis}
\end{tikzpicture}}
  \caption{Visualization of the $z = \sigma(x)\tanh(y)$ curve and the lower bound computed by linear programming. Red crosses represent the sampled points and dashed lines show the difference between the curve and the plane (summands in the optimization).}
  \label{fig:lp_bounds}
\vspace{-30pt}

\end{wrapfigure}

~\cref{fig:lp_bounds} shows an input region with Monte Carlo samples as red circles and summands in the LP objective as vertical lines. As this is a low-dimensional linear program (LP), we can solve it exactly in polynomial time using off-the-shelf LP solvers. We compute a candidate upper bound analogously.

\textbf{Step 2: Adjusting the offset to guarantee soundness}
Since we compute the lower bound from a subset of points in $B$, there can be a point in $B$ where the value of $h(x,y)$ is less than our computed lower bound. To ensure soundness, we compute $\Delta_l = \min_{(x, y) \in B} h(x, y) - (A_l \cdot x + B_l \cdot y + C_l)$ and then adjust the lower bound by updating the offset $C_l \leftarrow C_l + \Delta_l$, resulting in a sound lower bound plane. While the method of \cite{balunovic2019deepg} also performs offset calculation for obtaining sound bounds, they perform certification of image classifiers against geometric perturbations using expensive branch and bound for calculating the offset. In contrast, we exploit the structure of non-linearities used in LSTMs obtaining a closed-form formula for the offset yielding an exact solution. We now provide details of our offset adjustment method for $f(x,y) = \sigma(x) \tanh(y)$ and $g(x,y) = \sigma(x) y$.

\textbf{Offset calculation for $f(x, y) = \sigma(x) \tanh(y)$:}
Let $A_l \cdot x+B_l \cdot y+C_l$ be the initial lower bounding plane obtained from LP in region $B$. We define $F(x,y)$:
\[ F(x,y) = \sigma(x) \tanh(y) - (A_l \cdot x+B_l \cdot y+C_l) \]

To find $\Delta_l = \min_{(x,y) \in B} F(x,y)$, we first find the extreme points by computing partial derivatives.
\begin{align}
  \frac{\partial F}{\partial x} &= \sigma(x)\tanh(y)(1-\sigma(x)) - A_l \label{eq:dfdx} \\
  \frac{\partial F}{\partial y} &= \sigma(x)(1-\tanh^2(y)) - B_l \label{eq:dfdy}
\end{align}
We consider three cases:

\begin{itemize}
  \item \textit{Case 1: $x \in \{l_x,u_x\}$ and $y \in [l_y,u_y]$}
  Under this condition, we denote $S_x:=\sigma(x)$ as a constant.
  To ease notation, let $t = \tanh(y)$ where $t \in [\tanh(l_y), \tanh(u_y)]$.
  Then $\frac{\partial F}{\partial y} \mbeq 0$ can be rewritten as:
  \begin{equation} \label{eq:fixedx}
    1-t^2 = {B_l}/{S_x}
  \end{equation}

  \item \textit{Case 2: $y \in \{l_y,u_y\}$ and $x \in [l_x,u_x]$}
  Here we set $T_y:=\tanh(y)$ and $s = \sigma(x),\ x \in [\sigma(l_x), \sigma(u_x)]$ analogously.
  $\frac{\partial F}{\partial x} \mbeq 0$ is rewritten to:
  \begin{equation} \label{eq:fixedy}
    s(1-s) = {A_l}/{T_y}
  \end{equation}

  \item \textit{Case 3: otherwise}
  Otherwise, we consider both $\frac{\partial F}{\partial x} \mbeq 0$ and $\frac{\partial F}{\partial y} \mbeq 0$.
  By combining \cref{eq:dfdx} and \cref{eq:dfdy}, we reduce $\tanh(y)$ and obtain:
  \begin{equation} \label{eq:unfixed}
    s^4 + (-2-B_l)s^3 + (1+2B_l)s^2 + (-B_l)s - A_l^2 \mbeq 0
  \end{equation}
\end{itemize}

Given that $F(x,y)$ is differentiable and the region $B$ is compact, Fermat's theorem (stationary points) \cite{fermat_theorem} states that $F$ achieves its extremum at either the roots of \cref{eq:fixedx}, \cref{eq:fixedy}, and \cref{eq:unfixed}, or at the 4 corners of $B$. We evaluate $F$ at these points to get $\Delta_l$. We adjust the offset by replacing $C_l \leftarrow C_l + \Delta_l$. The adjusted $F$ is no less than 0 on any point in $B$, which means that the plane with adjusted $C_l$ becomes a sound lower bound of the $\sigma(x)\tanh(y)$ curve.

\textbf{Offset calculation for $g(x,y) = \sigma(x)y$:}
We next calculate the offset for $\sigma(x)y$. We define the differentiable function $G(x,y) = \sigma(x)y - (A_l \cdot x+B_l \cdot y+C_l)$ over the compact set $B$ and compute:

\begin{align}
  \frac{\partial G}{\partial x} &= \sigma(x)y(1-\sigma(x)) - A_l \label{eq:dfdx2} \\
  \frac{\partial G}{\partial y} &= \sigma(x) - B_l \label{eq:dfdy2}
\end{align}

We use Fermat's theorem and consider three cases:

\begin{itemize}
  \item \textit{Case 1: $x \in \{l_x,u_x\}$ and $y \in [l_y,u_y]$}
  When $\sigma(x)$ is fixed, \cref{eq:dfdy2} is constant, which means $G$ is monotonous in this case.

  \item \textit{Case 2: $y \in \{l_y,u_y\}$ and $x \in [l_x,u_x]$}
  Denote $s = \sigma(x)$ where $s \in [\sigma(l_x), \sigma(u_x)]$, then setting \cref{eq:dfdx2} $\mbeq 0$ becomes
  \begin{equation} \label{eq:fixedy2}
    s(1-s) = {A_l}/{y}
  \end{equation}

  \item \textit{Case 3: otherwise}
  If there is a local extremum in the region, the Hessian of $G$ must be either positive-definite or negative-definite.
  \begin{align*}
    &\frac{\partial^2 G}{\partial x^2} = \sigma(x)y(1-\sigma(x))(1-2\sigma(x)) ,
    \frac{\partial^2 G}{\partial y^2} = 0 ,
    \frac{\partial^2 G}{\partial x \partial y} = \sigma(x)(1-\sigma(x)) \\
    &\frac{\partial^2 G}{\partial x^2} \cdot \frac{\partial^2 G}{\partial y^2} - \left( \frac{\partial^2 G}{\partial x \partial y} \right)^2 = - \left( \sigma(x)(1-\sigma(x)) \right)^2 < 0
  \end{align*}
  Hence, there is no local extremum inside the boundaries.
\end{itemize}
To summarize, we only need to consider the roots of \cref{eq:fixedy2} to calculate the minimum of $G$ to obtain $\Delta_l$ for $\sigma(x)y$. \cref{fig:lp_bounds} shows the lower bound plane obtained after solving the LP and adjusting the offset. We update the upper bound analogously. 

\textbf{Asymptotic optimality}
We can prove that, similarly to \cite{balunovic2019deepg}, as we increase the number of samples $n$, the solution of the LP asymptotically approaches the solution of the original problem from \cref{eq:int_formulation}. Rephrasing and simplifying the theorem from \cite{balunovic2019deepg}:
\begin{theorem}
  Let $N$ be the number of points sampled in the algorithm. Let $(\bm{\omega}_l, b_l)$ be our lower constraint (linear constraints and bias, respectively) and let $L(\bm{\omega}^*, b^*)$ be the true minimum of function L. For every $\delta>0$ there exists $N_\delta$ such that $|L(\bm{\omega}_l, b_l)-L(\bm{\omega}^*, b^*)| < \delta$ for every $N > N_\delta$, with high probability. Analogous result holds for upper constraint $(\bm{\omega}_u, b_u)$ and function U.
\end{theorem}

We denote $L = \int_{(x,y)} F(x,y)$ and $(\bm{\omega}_l, b_l)$ are our $A_l, B_l, C_l$. Following the theorem, our sampling method guarantees the asymptotic optimality of our bounds. The theorem can be extended analogously for the upper bound.

\subsection{Abstraction refinement via optimization} \label{sec:opt_bounds}

While our approach based on sampling, linear programming, and Fermat's theorem allows us to obtain (asymptotically) optimal bounds, it still has a fundamental limitation that it produces a \emph{single} bound.
Further, this approach is, in a sense, greedy:
when considering the \emph{entire} network, it is possible that selecting non-optimal planes for each neuron yields more precise results at the end. Neither the method from \cite{ko2019popqorn} nor the method in \cref{sec:sampling_lp} achieves this. We present the first approach to learn an abstraction refinement that increases the end-to-end precision of certification.

\textbf{Step 1: Compute a set of candidate bounds}
We adapt our approach from \cref{sec:sampling_lp} to compute a set of candidate planes, instead of a single plane. We run the sampling procedure multiple times, each time on a different subregion of the original region $B = [l_x, u_x] \times [l_y, u_y]$, with the constraints still enforced over the entire region $B$. We define four different triangular subdomains: $\mathcal{T}_1$ and $\mathcal{T}_2$ are triangles resulting from splitting $B$ along the main diagonal, while $\mathcal{T}_3$ and $\mathcal{T}_4$ are triangles resulting from splitting $B$ along the other diagonal. We additionally define $\mathcal{T}_0 = B$. For each $k \in \{0, 1, 2, 3, 4\}$, we perform sampling and optimization as in~\cref{eq:lp_formulation2}, this time sampling from $\mathcal{T}_k$ to obtain candidate lower bounds:
\begin{align*}
  \min_{A_l, B_l, C_l \in \sR} & \sum_{i=1}^{n} \left( \sigma(x_i)\tanh(y_i) - (A_l \cdot x_i + B_l \cdot y_i + C_l) \right) \\
  \text{subject to } & \bigwedge_{i=1}^n A_l \cdot x_i + B_l \cdot y_i + C_l \leq \sigma(x_i)\tanh(y_i)
  \text{ where } (x_i, y_i) \sim \mathcal{T}_k
\end{align*}

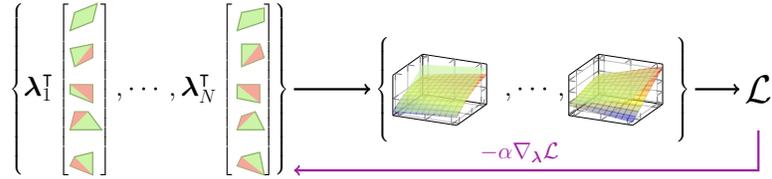
\begin{figure}[t]
  \centering
  \scalebox{0.6}{\definecolor{rightcol}{RGB}{198,243,158}
\definecolor{rightcolb}{RGB}{128,168,90}
\definecolor{lstmcol}{RGB}{255, 150, 150}
\definecolor{lambdacolb}{RGB}{159,34,163}

\begin{tikzpicture}[
  declare function={
    ub1(\x,\y) = 0.1133406783396896*\x+0.4609548699819045*\y-0.0022474648513164525;
    ub2(\x,\y) = 0.11679078744482975*\x+0.23723626586246435*\y+0.14235473664525983;
    curve(\x,\y) = tanh(\y) / (1 + exp(-\x));}]
  \node (lambda_layer) at (0, 0) {\LARGE $\left\{ \bm{\lambda}_1^\intercal \left[\begin{matrix} \hphantom{\cdots} \\ \hphantom{\cdots} \\ \hphantom{\cdots} \\ \hphantom{\cdots} \\ \hphantom{\cdots} \end{matrix}\right] , \cdots, \bm{\lambda}_N^\intercal \left[\begin{matrix} \hphantom{\cdots} \\ \hphantom{\cdots} \\ \hphantom{\cdots} \\ \hphantom{\cdots} \\ \hphantom{\cdots} \end{matrix}\right] \right\}$};

  \node (opt_layer) at ($(lambda_layer)+(8.5,0)$) {\LARGE $\left\{ \phantom{\bm{\lambda}_1^\intercal \left[\begin{matrix} \cdots \cdot \cdot \\ \cdots \\ \cdots \end{matrix}\right]} , \cdots, \phantom{\bm{\lambda}_N^\intercal \left[\begin{matrix} \cdots \cdot \\ \cdots \\ \cdots \end{matrix}\right]} \right\}$};

  \node (loss_layer) at ($(opt_layer)+(5.0,0)$) {\Huge $\gL$};

  \node (update_text) at ($(lambda_layer.east)+(5.0,-1.4)$) [text=lambdacolb] {\Large $-\lr \nabla_{\bm{\lambda}} \gL$};

  \draw[->, line width=0.5mm] (lambda_layer) -- (opt_layer);
  \draw[->, line width=0.5mm] (opt_layer) -- (loss_layer);
  \draw[->, line width=0.5mm, draw=lambdacolb] ($(loss_layer.south)+(0,-0.5)$) |- ($(lambda_layer.east)+(0,-1.8)$);
  
  \coordinate (lmb1ctr) at ($(lambda_layer)+(-1.45, 0)$);
  \coordinate (lmb10) at ($(lmb1ctr)+(0, 1.6)$);
  \draw[fill=rightcol, draw=rightcolb, opacity=0.9, thick]
    ($(lmb10)+(0.3,0.3)$) --
    ($(lmb10)+(-0.2,0.1)$) --
    ($(lmb10)+(-0.3,-0.3)$) --
    ($(lmb10)+(0.2,-0.1)$) --cycle;

  \coordinate (lmb11) at ($(lmb1ctr)+(0, 0.8)$);
  \draw[fill=rightcol, draw=rightcolb, opacity=0.9, thick]
    ($(lmb11)+(0.2,0.2)$) --
    ($(lmb11)+(-0.3,0.1)$) --
    ($(lmb11)+(-0.2,-0.3)$) --
    ($(lmb11)+(0.2,-0.1)$) -- cycle;
  \draw[fill=lstmcol, draw=none, opacity=0.8, thick]
    ($(lmb11)+(0.2,0.2)$) --
    ($(lmb11)+(0.2,-0.1)$) --
    ($(lmb11)+(-0.2,-0.3)$) -- cycle;

  \coordinate (lmb12) at ($(lmb1ctr)+(0, 0)$);
  \draw[fill=rightcol, draw=rightcolb, opacity=0.9, thick]
    ($(lmb12)+(-0.3,0.1)$) --
    ($(lmb12)+(-0.3,-0.1)$) --
    ($(lmb12)+(0.2,-0.3)$) --
    ($(lmb12)+(0.2,0.1)$) -- cycle;
  \draw[fill=lstmcol, draw=none, opacity=0.8, thick]
    ($(lmb12)+(-0.3,0.1)$) --
    ($(lmb12)+(0.2,0.1)$) --
    ($(lmb12)+(0.2,-0.3)$) -- cycle;

  \coordinate (lmb13) at ($(lmb1ctr)+(0, -0.8)$);
  \draw[fill=rightcol, draw=rightcolb, opacity=0.9, thick]
    ($(lmb13)+(-0.3,-0.1)$) --
    ($(lmb13)+(0.4,-0.1)$) --
    ($(lmb13)+(0.1,0.3)$) --
    ($(lmb13)+(-0.2,0.3)$) -- cycle;
  \draw[fill=lstmcol, draw=none, opacity=0.8, thick]
    ($(lmb13)+(-0.3,-0.1)$) --
    ($(lmb13)+(-0.2,0.3)$) --
    ($(lmb13)+(0.1,0.3)$) -- cycle;

  \coordinate (lmb14) at ($(lmb1ctr)+(0, -1.6)$);
  \draw[fill=rightcol, draw=rightcolb, opacity=0.9, thick]
    ($(lmb14)+(0.2,-0.3)$) --
    ($(lmb14)+(0.2,0.2)$) --
    ($(lmb14)+(-0.2,0.1)$) --
    ($(lmb14)+(-0.3,-0.1)$) -- cycle;
  \draw[fill=lstmcol, draw=none, opacity=0.8, thick]
    ($(lmb14)+(0.2,-0.3)$) --
    ($(lmb14)+(-0.3,-0.1)$) --
    ($(lmb14)+(-0.2,0.1)$) -- cycle;

  \coordinate (lmb2ctr) at ($(lambda_layer)+(2.25, 0)$);
  \coordinate (lmb20) at ($(lmb2ctr)+(0, 1.6)$);
  \draw[fill=rightcol, draw=rightcolb, opacity=0.9, thick]
    ($(lmb20)+(0.3,0.2)$) --
    ($(lmb20)+(-0.2,0.1)$) --
    ($(lmb20)+(-0.3,-0.3)$) --
    ($(lmb20)+(0.3,-0.1)$) --cycle;

  \coordinate (lmb21) at ($(lmb2ctr)+(0, 0.8)$);
  \draw[fill=rightcol, draw=rightcolb, opacity=0.9, thick]
    ($(lmb21)+(0.2,0.2)$) --
    ($(lmb21)+(-0.2,0.2)$) --
    ($(lmb21)+(-0.2,-0.3)$) --
    ($(lmb21)+(0.3,-0.1)$) -- cycle;
  \draw[fill=lstmcol, draw=none, opacity=0.8, thick]
    ($(lmb21)+(0.2,0.2)$) --
    ($(lmb21)+(0.3,-0.1)$) --
    ($(lmb21)+(-0.2,-0.3)$) -- cycle;

  \coordinate (lmb22) at ($(lmb2ctr)+(0, 0)$);
  \draw[fill=rightcol, draw=rightcolb, opacity=0.9, thick]
    ($(lmb22)+(-0.3,0.1)$) --
    ($(lmb22)+(-0.3,-0.2)$) --
    ($(lmb22)+(0.2,-0.3)$) --
    ($(lmb22)+(0.2,0.1)$) -- cycle;
  \draw[fill=lstmcol, draw=none, opacity=0.8, thick]
    ($(lmb22)+(-0.3,0.1)$) --
    ($(lmb22)+(0.2,0.1)$) --
    ($(lmb22)+(0.2,-0.3)$) -- cycle;

  \coordinate (lmb23) at ($(lmb2ctr)+(0, -0.8)$);
  \draw[fill=rightcol, draw=rightcolb, opacity=0.9, thick]
    ($(lmb23)+(-0.3,-0.1)$) --
    ($(lmb23)+(0.3,-0.1)$) --
    ($(lmb23)+(0.1,0.3)$) --
    ($(lmb23)+(-0.2,0.2)$) -- cycle;
  \draw[fill=lstmcol, draw=none, opacity=0.8, thick]
    ($(lmb23)+(-0.3,-0.1)$) --
    ($(lmb23)+(-0.2,0.2)$) --
    ($(lmb23)+(0.1,0.3)$) -- cycle;

  \coordinate (lmb24) at ($(lmb2ctr)+(0, -1.6)$);
  \draw[fill=rightcol, draw=rightcolb, opacity=0.9, thick]
    ($(lmb24)+(0.3,-0.3)$) --
    ($(lmb24)+(0.2,0.2)$) --
    ($(lmb24)+(-0.2,0.1)$) --
    ($(lmb24)+(-0.3,-0.1)$) -- cycle;
  \draw[fill=lstmcol, draw=none, opacity=0.8, thick]
    ($(lmb24)+(0.3,-0.3)$) --
    ($(lmb24)+(-0.3,-0.1)$) --
    ($(lmb24)+(-0.2,0.1)$) -- cycle;

  \coordinate (opt1) at ($(opt_layer)+(-3.1,-0.8)$);
  \begin{axis}[width=0.30\linewidth,at=(opt1),
    domain=0.4:1.6, domain y=0.4:1.6, grid=both,
    xlabel=\empty,xticklabels={,,},
    ylabel=\empty,yticklabels={,,},
    zlabel=\empty,zticklabels={,,},
    zlabel style={rotate=-90}]
    \addplot3[surf, samples=12, opacity=0.5] {curve(x,y)};
    \addplot3[surf, samples=2, opacity=0.01, fill opacity=0.6, color=rightcol, faceted color=rightcol] {ub1(x,y)};
  \end{axis}

  \coordinate (opt2) at ($(opt_layer)+(0.8,-0.8)$);
  \begin{axis}[width=0.30\linewidth,at=(opt2),
     domain=-1.2:1.2, domain y=-0.22:0.64, grid=both,
     xlabel=\empty,xticklabels={,,},
     ylabel=\empty,yticklabels={,,},
     zlabel=\empty,zticklabels={,,},
     zlabel style={rotate=-90}]
     \addplot3[surf, samples=12, opacity=0.5] {curve(x,y)};
     \addplot3[surf, samples=2, opacity=0.01, fill opacity=0.6, color=rightcol, faceted color=rightcol] {ub2(x,y)};
   \end{axis}

\end{tikzpicture}}
  \caption{Learning to combine linear bounds via gradient descent. Here the five candidate planes multiplied by $\bm{\lambda}$ are depicted either in green or red, or both. Green represents the sampled domain, $\mathcal{T}_k$, and red is the extension of the obtained green plane out of the domain. With the linear combination of the planes, we compute the bound, calculate the loss, and backpropagate.}
  \label{fig:opt_method}
\end{figure}

For each neuron $i$, this yields 5 candidate lower bound and upper bound planes, $LB^k_{i}$ and $UB^k_{i}$ for $k \in \{0, 1, 2, 3, 4\}$. These five candidate planes for each of the $N$ neurons are shown in~\cref{fig:opt_method}.

\textbf{Step 2: Find the optimal combinations of the bounds}
Next, our goal is to learn a linear combination of the computed candidate bounds which yields the highest end-to-end certification precision for the given input region. To do this, we define the lower and upper bound of neuron $i$ as a linear combination of the proposed five bounds:
\begin{align*}
  LB_i = \sum_{k=0}^4 \lambda^{LB}_i \cdot LB_i^k,\ \ \sum_{k=0}^4 \lambda^{LB}_i = 1,
  UB_i = \sum_{k=0}^4 \lambda^{UB}_i \cdot UB_i^k,\ \ \sum_{k=0}^4 \lambda^{UB}_i = 1.
\end{align*}

Recall that we formulate robustness certification as proving that for all labels $i$ different from the ground truth label $t$: $z_t - z_i > 0$. The lower bound on $z_t - z_i$ is computed using backsubstitution \cite{singh2019abstract}, as shown in our overview example in~\cref{sec:overview}. However, this lower bound now depends on the coefficients $\bm{\lambda}$, so we define the function $f(x, \epsilon, i, \bm{\lambda})$ which computes the lower bound of the expression $z_t - z_i$ when using $\bm{\lambda}$ to combine the neuron bounds.

We describe our approach to find the best coefficients $\bm{\lambda}$ in \cref{alg:optbound}. Consider the number of possible labels $m$ and the number of binary operations of interest $N_{ops}$. To find $\bm{\lambda}$, we solve the optimization problem for each label $i$:
\begin{equation*}
  z_t - z_i > \max_{\bm{\lambda}} f(x, \epsilon, i, \bm{\lambda})
\end{equation*}
If the solution to the above optimization problem is positive, then we proved robustness with respect to class $i$. In the algorithm, we initialize $\tilde{\bm{\lambda}}$, a pre-normalized vector of $\bm{\lambda}$, for each neuron, uniformly between -1 and 1. Then, in each epoch, we compute the normalized $\bm{\lambda}$ by applying softmax to $\tilde{\bm{\lambda}}$ and run certification using $\bm{\lambda}$, obtaining a loss $\gL$ equal to the value $-f(x, \epsilon, i, \bm{\lambda})$. We perform gradient descent update on $\tilde{\bm{\lambda}}$ based on the loss. If the loss is negative, we have found $\bm{\lambda}$ which proves the robustness and the algorithm terminates. The core updating flow is shown in~\cref{fig:opt_method}.

\begin{algorithm}[t]
  \caption{Learning $\bm{\lambda}$ via gradient descent}
  \label{alg:optbound}
  \begin{algorithmic}
    \STATE {\bfseries Given} input $\vx$, label $y$, model $\gM$, perturbation $\epsilon$
    \STATE Initialize the polyhedral abstractions and candidate bounds based on $\vx$, $\gM$ and $\epsilon$.
    \FOR{$i \leftarrow 1$ {\bfseries to} $m$ {\bfseries where} $i \neq y$}
      \STATE Initialize $\tilde{\bm{\lambda}} \sim [-1, 1]^{N_{ops} \times 5}$,
     $epoch \leftarrow 0$
      \REPEAT
        \STATE $\bm{\lambda} \leftarrow \mathrm{SoftMax}(\tilde{\bm{\lambda}})$,
	  $\gL \leftarrow -f(x, \epsilon, i, \bm{\lambda})$,
	$\tilde{\bm{\lambda}} \leftarrow \tilde{\bm{\lambda}} - \lr \nabla_{\tilde{\bm{\lambda}}} \gL$,
	 $epoch \leftarrow epoch + 1$
      \UNTIL{$epoch = max\_epoch$ or $\gL < 0$}
      \IF{$\gL \geq 0$}
        \STATE {\bfseries return} not certified
      \ENDIF
    \ENDFOR
    \STATE {\bfseries return} certified
  \end{algorithmic}
\end{algorithm}

\section{Certification of Speech Preprocessing} \label{sec:spprep}

Speech preprocessing transforms the original set of perturbed speech signals, represented via intervals, through complex pipeline operations, into a non-linear and non-convex set. Propagating this set through the network is computationally expensive (infeasible for large models). To address this issue, we define precise overapproximations of key non-linear operations found in the speech preprocessing pipeline, such as Square and Log, expressed in the \deeppoly abstraction. These approximate bounds are computed via constant time closed form formulas based on concrete bounds of the inputs. We note that the first and third stages of the pipeline described in \cref{sec:mel_explained} involve an affine transformation, captured exactly using \deeppoly. Overall, when combined with our LSTM verifier, this method yields more precise end-to-end certification results than using intervals for approximating speech preprocessing.

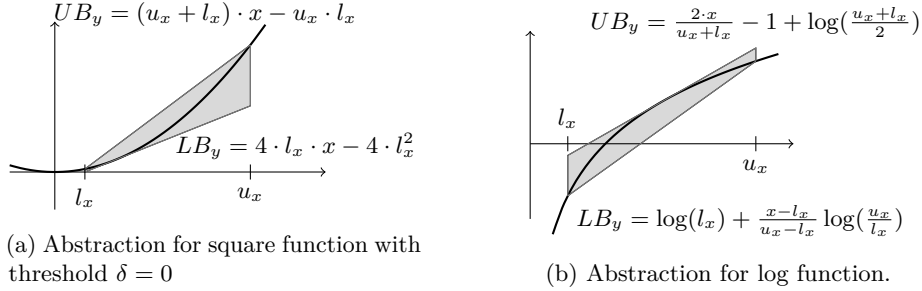
\begin{figure}[t]
  \centering
  \begin{subfigure}[b]{0.45\textwidth}
    \centering
    \scalebox{1.0}{\begin{tikzpicture}[xscale=2]
  \begin{scope}[xshift=0cm]
    \draw[fill=gray!30] (0.2, 0)-- (1.3, 0.88)  -- (1.3, 1.69) -- (0.2, 0.04)  -- cycle;
    \draw[->] (0, -0.5) -- (0, 2) node[anchor=east] {};
    \draw[->] (-0.3, 0) -- (1.8, 0) node[anchor=north] {};
    \draw[thick, domain=-0.3:1.4, smooth, variable=\x] plot ({\x}, {\x*\x});
    \draw[gray, domain=0.2:1.3, variable=\x] plot ({\x}, {-0.2*1.3+\x*1.5});
    \node[text=black] at (1.0, 2.1) {\small $UB_y = (u_x + l_x) \cdot x - u_x \cdot l_x$};
    \draw[gray, domain=0.2:1.3, variable=\x] plot ({\x}, {0.8*\x-0.16});
    \node[text=black] at (1.6, 0.35) {\small $LB_y= 4 \cdot l_x \cdot x -4 \cdot l_x^2$};
    \draw[gray, domain=0:0.04, variable=\y] plot ({0.2}, {\y});
    \draw[gray, domain=0.88:1.69, variable=\y] plot ({1.3}, {\y});
    \draw[domain=0.1:-0.1, variable=\y] plot ({0.2}, {\y}) node[below] {$l_x$};
    \draw[domain=0.1:-0.1, variable=\y] plot ({1.3}, {\y})  node[below] {$u_x$};
  \end{scope}
\end{tikzpicture}}
    \caption{Abstraction for square function with threshold $\delta=0$}
    \label{fig:square}
  \end{subfigure}
	\hfill
  \begin{subfigure}[b]{0.45\textwidth}
    \centering
    \scalebox{1.0}{\begin{tikzpicture}
  \begin{scope}[xshift=4cm, yshift=0.5cm]
    \draw[fill=gray!30] (0.5, {-2.5/3.5+ln(3.5/2)})-- (0.5, {-ln(2)})  -- (3, {ln(0.5)+ln(6)}) -- (3, {2.5/3.5+ln(1.75)})  -- cycle;
    \draw[->] (0, -1) -- (0, 1.5) node[anchor=east] {};
    \draw[->] (0, 0) -- (3.5, 0) node[anchor=north] {};
    \draw[thick, domain=0.3:3.3, smooth, variable=\x] plot ({\x}, {ln(\x)});
    \draw[gray, domain=3:0.5, variable=\x] plot ({\x}, {ln(0.5)+(\x-0.5)/2.5*ln(6)})
    node[below right, text=black] {\small $LB_y = \log (l_x) + \frac{x-l_x}{u_x-l_x} \log (\frac{u_x}{l_x})$};
    \draw[gray, domain=0.5:3, variable=\x] plot ({\x}, {2*\x/3.5-1+ln(3.5/2)})
    node[above, text=black] {\small $UB_y = \frac{2\cdot x}{u_x+l_x} -1 + \log (\frac{u_x+l_x}{2})$};
    \draw[gray, domain=-0.693:-0.155, variable=\y] plot({0.5}, {\y});
    \draw[gray, domain=1.099:1.274, variable=\y] plot({3}, {\y});
    \draw[domain=-0.1:0.1, variable=\y] plot ({0.5}, {\y}) node[above] {$l_x$};
    \draw[domain=0.1:-0.1, variable=\y] plot ({3}, {\y}) node[below] {$u_x$};
    
  \end{scope}
\end{tikzpicture}}
    \caption{Abstraction for log function.}
    \label{fig:log}
  \end{subfigure}
  \caption{Two polyhedral abstractions for the speech preprocessing stage.}
  \label{fig:unary_relaxation}
\vspace{-0.7cm}
\end{figure}

\textbf{Square}
The lower and upper polyhedral bounds of the output of the square function $y = x^2$ where $x \in [l_x, u_x]$ are shown in~\cref{fig:square}. We first consider the bounds for $y$ which minimize the area in the $xy$-plane. The upper bound $UB_y$ is obtained by computing the chord joining the end points $(l_x,l_x^2)$ and $(u_x,u_x^2)$. The lower bound is a line parallel to $UB_y$ passing through a point $((u_x + l_x) / 2, ((u_x + l_x) / 2)^2)$ in the middle of the curve.
\begin{align*}
  LB_y &= (u_x + l_x) \cdot x - \left( (u_x + l_x) / 2 \right)^2,
  UB_y = (u_x + l_x) \cdot x - u_x \cdot l_x.
\end{align*}
While the above bounds would be sufficient in any other domain, they do not work for the speech domain as the subsequent Log requires that the input is strictly non-negative, as it is not defined for negative inputs. Also, we should carefully consider the floating point errors during calculations. Hence, we introduce the additional parameter $\delta \in \mathbb{R}$, a small threshold value to ensure the lower bound stays non-negative. In our experiments, we set $\delta = 1 \times 10^{-5}$. Upper and lower bounds for $y = x^2$ are computed as $UB_y = (u_x + l_x) \cdot x - u_x \cdot l_x$ and $LB_y=$
\begin{align*}
  \begin{cases}
    2 \cdot (l_x+\sqrt{l_x^2 - \delta})  \cdot x  - (l_x+\sqrt{l_x^2 -\delta})^2 & 3 \cdot l_x^2 + 2\cdot l_x . u_x - u_x^2  \leq 4 \cdot \delta, \sqrt{\delta} \leq l_x\\
    2 \cdot (u_x - \sqrt{u_x^2 - \delta}) \cdot  x - (u_x - \sqrt{u_x^2 - \delta})^2 & 3 \cdot u_x^2 + 2 . u_x . l_x - l_x^2  \leq 4 \cdot \delta, u_x \leq -\delta \\
    0 & l_x \leq \sqrt{\delta}, -\sqrt{\delta} \leq u_x \\
    (u_x + l_x) \cdot x - \left( (u_x + l_x) / 2 \right)^2 & o.w. \\
  \end{cases}
\end{align*}

\textbf{Log}
We define the polyhedral abstraction of the output $y = \log (x)$ of the log operation where $x \in [l_x, u_x]$, as shown in ~\cref{fig:log}. Our abstractions are optimal and minimize the area in the $xy$-plane. The lower bound $LB_y$ is the chord joining the end points $(l_x, \log(l_x))$ and $(u_x, \log(u_x))$. The upper bound $UB_y$ is obtained by computing a line parallel to $LB_y$ passing through the middle of the curve at $((u_x + l_x) / 2, \log((u_x + l_x) / 2))$. Our final abstractions are:
\begin{align*}
  LB_y = \log (l_x) + \frac{x-l_x}{u_x-l_x} \log (\frac{u_x}{l_x}),
  UB_y = \frac{2\cdot x}{u_x+l_x} -1 + \log (\frac{u_x+l_x}{2}).
\end{align*}

\section{Experimental Evaluation} \label{sec:experiments}

We implemented our approach in a verifier called \tool, using PyTorch \cite{paszke2017automatic} and Gurobi 9.0 for solving linear programs. The code is available in \url{https://github.com/eth-sri/prover}. We evaluate \tool on speech classifiers for FSDD \cite{zohar2020free} and GSC v2 \cite{warden2018speech} datasets. Then, we compare \tool against POPQORN \cite{ko2019popqorn} on the MNIST image classification task proposed by it. We note that POPQORN does not scale to the speech classifiers considered in our work. We demonstrate further scalability by verifying large motion sensor sequence classifier trained on HAPT \cite{reyes2016transition} dataset containing 256 hidden dimensional 4 layered LSTM units.

\textbf{Setup}
GSC dataset experiments ran on an Nvidia GeForce RTX 2080, while the rest ran on a single Tesla V100. Following convention from prior work \cite{singh2019abstract}, we consider only those inputs that are classified correctly without perturbation. We use the same set of hyperparameters for the experiments unless specifically mentioned. We use $100$ sampling points for constructing the linear program and optimize $\bm{\lambda}$ parameters using Adam \cite{kingma2014adam} for $100$ epochs. During optimization, we initialize the learning rate to $100$ and multiply it by $0.98$ after every epoch.

\subsection{Speech classification} \label{sec:exp_speech}

We certify the robustness of two speech classifiers for the FSDD and GSC v2 datasets. FSDD consists of recordings of digits spoken by six different speakers, recorded at 8kHz. GSC has 35 distinct labels of single word utterances at 16kHz. We compare our base method based on sampling and linear programming (\cref{sec:sampling_lp}), denoted as \tool(LP), and our method using abstraction refinement via optimization  (\cref{sec:opt_bounds}), denoted as \tool(OPT).

\begin{figure}[t]
  \centering
  \begin{subfigure}{0.45\textwidth}
    \centering
    \scalebox{0.6}{\begin{tikzpicture}
  \begin{axis}[
    grid style={draw=white},
    ymajorgrids=true,
    xmin=-95, xmax=-65,
    ymin=0,    ymax=105,
    minor tick num=0,
    axis x line=bottom,
    axis y line=left,
    legend style={fill=none, draw=none, at={(0.05, 0.35)}, anchor=north west},
    legend style={fill=none, draw=none},
    legend cell align={left},
    xlabel={\Large $dB_{\bm{s}}(\bm{\delta})$},
    ylabel={\Large Certified(\%)},
    xtick=data,
    every axis y label/.style={
      at={(0.5, 1.0)}, above=1mm
    },
    axis background/.style={fill=gray!15}
    ]

    \addplot+ [draw=OliveGreen, mark=*, mark options={scale=1, fill=OliveGreen}]
    table [x=x, y=precision, col sep=comma]{
    x,  precision
      -90,  90
      -80,  80
      -70,  58
    };

    \addplot+ [draw=NavyBlue, mark=*, mark options={scale=1, fill=NavyBlue}]
    table [x=x, y=precision, col sep=comma]{
      x,  precision
      -90,  86
      -80,  70
      -70,  25
    };

    \legend{
      \large \tool(OPT),
      \large \tool(LP)};

  \end{axis}
\end{tikzpicture}}
    \caption{FSDD precision}
    \label{fig:fsdd_precision}
  \end{subfigure}
  \begin{subfigure}{0.45\textwidth}
    \centering
    \scalebox{0.6}{\begin{tikzpicture}
  \begin{axis}[
    grid style={draw=white},
    ymajorgrids=true,
    xmin=-95, xmax=-65,
    ymin=0,    ymax=550,
    minor tick num=0,
    axis x line=bottom,
    axis y line=left,
    legend style={fill=none, draw=none, at={(0.05, 0.9)}, anchor=north west},
    legend style={fill=none, draw=none},
    legend cell align={left},
    xlabel={\Large $dB_{\bm{s}}(\bm{\delta})$},
    ylabel={\Large Running time(s)},
    xtick=data,
    every axis y label/.style={
      at={(0.5, 1.0)}, above=1mm
    },
    axis background/.style={fill=gray!15}
    ]

    \addplot+ [draw=OliveGreen, mark=*, mark options={scale=1, fill=OliveGreen}]
    table [x=x, y=runtime, col sep=comma]{
      x,  runtime
      -90,  218
      -80,  293
      -70,  485
    };

    \addplot+ [draw=NavyBlue, mark=*, mark options={scale=1, fill=NavyBlue}]
    table [x=x, y=runtime, col sep=comma]{
      x,  runtime
      -90,  41.5551
      -80,  41.4420
      -70,  40.8685
    };

    \legend{
      \large \tool(OPT),
      \large \tool(LP)};

  \end{axis}
\end{tikzpicture}}
    \caption{FSDD runtime}
    \label{fig:fsdd_runtime}
  \end{subfigure}
  \begin{subfigure}{0.45\textwidth}
    \centering
    \scalebox{0.6}{\begin{tikzpicture}
  \begin{axis}[
    grid style={draw=white},
    ymajorgrids=true,
    xmin=-115, xmax=-85,
    ymin=0,    ymax=80,
    minor tick num=0,
    axis x line=bottom,
    axis y line=left,
    legend style={fill=none, draw=none, at={(0.05, 0.35)}, anchor=north west},
    legend style={fill=none, draw=none},
    legend cell align={left},
    xlabel={\Large $dB_{\bm{s}}(\bm{\delta})$},
    ylabel={\Large Certified(\%)},
    xtick=data,
    every axis y label/.style={
      at={(0.5, 1.0)}, above=1mm
    },
    axis background/.style={fill=gray!15}
    ]

    \addplot+ [draw=OliveGreen, mark=*, mark options={scale=1, fill=OliveGreen}]
    table [x=x, y=precision, col sep=comma]{
    x,  precision
      -110,  75
      -100,  46
      -90,  25
    };

    \addplot+ [draw=NavyBlue, mark=*, mark options={scale=1, fill=NavyBlue}]
    table [x=x, y=precision, col sep=comma]{
      x,  precision
      -110,  70
      -100,  37
      -90,  15
    };

    \legend{
      \large \tool(OPT),
      \large \tool(LP)};

  \end{axis}
\end{tikzpicture}}
    \caption{GSC precision}
    \label{fig:gsc_precision}
  \end{subfigure}
  \begin{subfigure}{0.45\textwidth}
    \centering
    \scalebox{0.6}{\begin{tikzpicture}
  \begin{axis}[
    grid style={draw=white},
    ymajorgrids=true,
    xmin=-115, xmax=-85,
    ymin=0,    ymax=800,
    minor tick num=0,
    axis x line=bottom,
    axis y line=left,
    legend style={fill=none, draw=none, at={(0.05, 0.9)}, anchor=north west},
    legend style={fill=none, draw=none},
    legend cell align={left},
    xlabel={\Large $dB_{\bm{s}}(\bm{\delta})$},
    ylabel={\Large Running time(s)},
    xtick=data,
    every axis y label/.style={
      at={(0.5, 1.0)}, above=1mm
    },
    axis background/.style={fill=gray!15}
    ]

    \addplot+ [draw=OliveGreen, mark=*, mark options={scale=1, fill=OliveGreen}]
    table [x=x, y=runtime, col sep=comma]{
      x,  runtime
      -110,  403.1839550471306
      -100,  448.3649878668785
      -90,  728.8733764100075
    };

    \addplot+ [draw=NavyBlue, mark=*, mark options={scale=1, fill=NavyBlue}]
    table [x=x, y=runtime, col sep=comma]{
      x,  runtime
      -110,  24.9613
      -100,  27.8979
      -90,  26.5885
    };

    \legend{
      \large \tool(OPT),
      \large \tool(LP)};

  \end{axis}
\end{tikzpicture}}
    \caption{GSC runtime}
    \label{fig:gsc_runtime}
  \end{subfigure}
  \caption{
    Performance plots for the FSDD and GSC datasets with different perturbations.
    All tests are done with the same architecture described in the text.
  }
  \label{fig:fsdd_plot}
  \vspace{-0.7cm}
\end{figure}
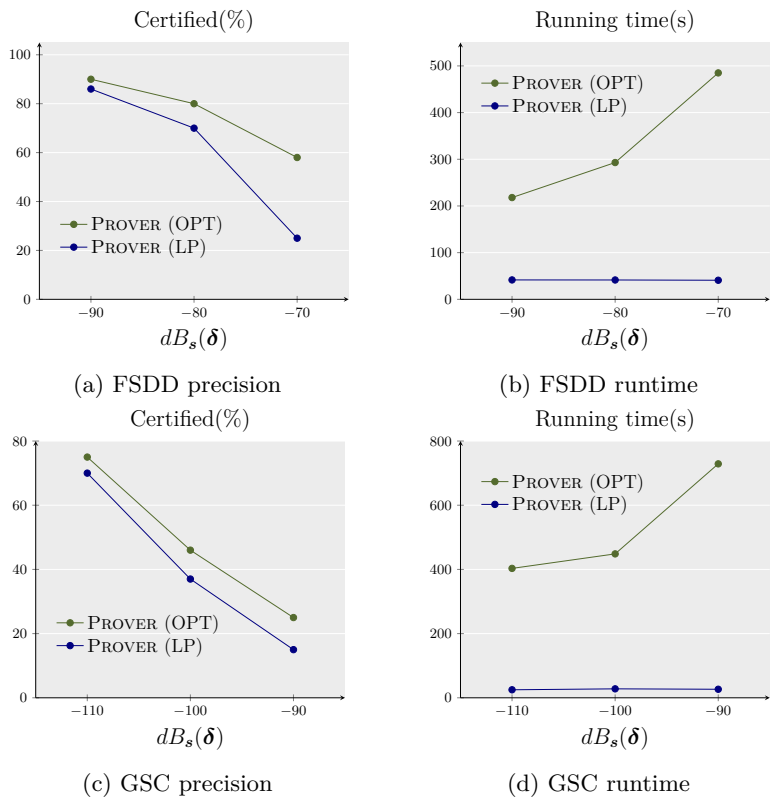

\textbf{Preprocessing}
A key challenge in speech classification, not encountered in the image domain, is the complex preprocessing stage before the LSTM network. The preprocessing stage in this experiment consists of FFT and Mel-filter transformations. Preprocessed input then passes through the fully connected layer with ReLU activation followed by the LSTM unit.

\textbf{FSDD certification}
We used the following parameters for the preprocessing: we slice the raw wave signal with length 256 using a stride of 200 with 10 Mel-frequencies. For this experiment, we trained an LSTM network with two LSTM layers and 32 hidden units each,
preceded by a 40 ReLU-activated fully-connected layer. This network achieves an accuracy of 83.6\% on the FSDD task.
The average number of frames was 14.7. We verify the first 100 correctly classified inputs for each perturbation.
Our perturbation metric on speech classification tasks is described in \cref{sec:threat_model}. Our results are shown in~\cref{fig:fsdd_precision} and~\cref{fig:fsdd_runtime}. We vary the decibel perturbation between -90 dB and -70 dB and evaluate the precision and runtime of \tool. \cref{fig:fsdd_precision} shows the percentage of certified samples: our method based on optimizing the bounds (OPT) performs best, \eg certifying twice as many samples compared to LP,  for a significant perturbation of -70 dB. In terms of runtime, \cref{fig:fsdd_runtime} shows that the OPT runtime increases with the perturbation magnitude, meaning that the optimizer needs more iterations to converge to the resulting bounds.

\textbf{Interval vs. Polyhedral abstraction for speech preprocessing}
We studied experimentally the importance of designing precise polyhedral abstractions of the speech preprocessing pipeline. If we replace the polyhedral bounds for the square and logarithm operations with interval constraints, the precision of \tool(LP) drops from 86\% to 61\% on -90dB and from 70\% to 20\% on -80dB. This shows the importance of keeping relational information while overapproximating the speech preprocessing pipeline.

\textbf{GSC certification}
We used the following parameters for the preprocessing: we downsample the raw input to 8kHz, sliced the signal in length 1024, followed by 10 Mel-frequency filterbanks. As with the FSDD architecture, we used two layers of LSTM and 50 hidden units each, preceded by a 50 ReLU-activated fully-connected layer. This network achieves accuracy of 80\% on the GSC task.
Certifying the GSC classifier is more challenging than FSDD: this dataset has 35 labels, compared to 10 in FSDD. The larger label set size requires \tool to compare 34 output differences - acquiring the lower bounds of $l_{GT} - l_{FL}$ where each term stands for the final output score for the ground truth and false label, respectively.
\cref{fig:gsc_precision} shows the percentage of certified samples: 75\% on -110dB and 46\% on -100dB with \tool(OPT), again higher precision than \tool(LP). \cref{fig:gsc_runtime} shows the longer running time for \tool(OPT) than on FSDD, due to its larger label set size.

\subsection{Image classification}

Based on the setup from \cite{ko2019popqorn}, we flatten each image into a vector of dimension 784. This vector is partitioned into a sequence of $f$ frames ($f$ depends on the experiment). Next, the LSTM uses this frame as an input.

\textbf{Comparison with POPQORN}
We compare the precision and scalability of \tool against POPQORN \cite{ko2019popqorn}. We trained an LSTM network containing 1 layer with 32 hidden units using standard training, achieving an accuracy of 96.5\%. The network receives a sequence of $f = 7$ image slices as input and predicts a digit corresponding to the image.

As POPQORN is slow, we used only ten correctly classified images randomly sampled from the test set. For each frame index $i$ and each method, we compute the maximum perturbation bound $\epsilon$ such that the method can certify that the LSTM classifier is robust to perturbations up to $\epsilon$ in the $L_\infty$-norm of the $i$-th slice of the image. We determine the maximum $\epsilon$ using the same binary search procedure as in \cite{ko2019popqorn}.

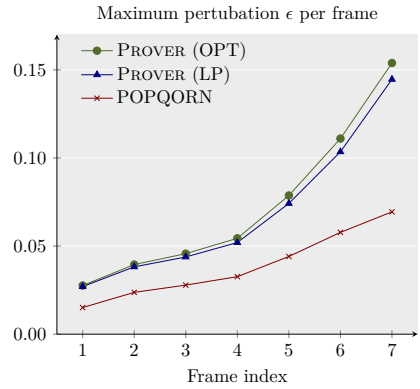
\begin{wrapfigure}{r}{0.45\textwidth}
  \vspace{-20pt}
    \centering
    \scalebox{0.7}{\begin{tikzpicture}
  \begin{axis}[
    grid style={draw=white},
    ymajorgrids=true,
    xmin=0.5, xmax=7.5,
    ymin=0,    ymax=0.17,
    minor tick num=0,
    axis x line=bottom,
    axis y line=left,
    legend style={fill=none, draw=none, at={(0.05, 1.0)}, anchor=north west},
    legend style={fill=none, draw=none},
    legend cell align={left},
    xlabel={Frame index},
    ylabel={Maximum pertubation $\epsilon$ per frame},
    xtick=data,
    every axis y label/.style={
          at={(0.5, 1.0)}, above=1mm
    },
    y tick label style={
          /pgf/number format/.cd,
          fixed,
          fixed zerofill,
          precision=2,
          /tikz/.cd,
    },
    axis background/.style={fill=gray!15}
    ]

    \addplot+ [draw=OliveGreen, mark=*, mark options={scale=1, fill=OliveGreen}]
    table [x=x, y=eps, col sep=comma]{
      x,  eps
      1,  0.0276
      2,  0.0395
      3,  0.0457
      4,  0.0544
      5,  0.0787
      6,  0.1110
      7,  0.1539
    };

    \addplot+ [draw=NavyBlue, mark=triangle*, mark options={scale=1, fill=NavyBlue}]
    table [x=x, y=eps, col sep=comma]{
      x,  eps
      1,  0.0270
      2,  0.0382
      3,  0.0438
      4,  0.0520
      5,  0.0742
      6,  0.1035
      7,  0.1446
    };

    \addplot+ [draw=darkred, mark=x, mark options={scale=1, fill=darkred}]
    table [x=x, y=eps, col sep=comma]{
      x,  eps
      1,  0.01511
      2,  0.02373
      3,  0.02785
      4,  0.03263
      5,  0.04410
      6,  0.05776
      7,  0.06942
    };

    \legend{\tool(OPT), \tool(LP), POPQORN};

  \end{axis}
\end{tikzpicture}}
    \caption{
      Results for the comparison between \tool and POPQORN. Plotted points represent the maximum $L_\infty$ norm perturbation for each frame index 1 through 7.
    }
    \label{fig:popqorn_comparison}
  \vspace{-20pt}
\end{wrapfigure}

~\cref{fig:popqorn_comparison} presents the results of this experiment. We observe that, for all three methods, early frames have smaller $\epsilon$ certified perturbation bounds than the later frames. The reason is that the approximation error on frame 1 propagates through the later frames to the classifying layer, while the error on frame 7 only affects the last layer. Across all frames, both our LP and OPT methods significantly outperform POPQORN, meaning that \tool enables a more precise abstraction than POPQORN. As for speech classifiers, OPT is more precise than LP.
We compare running times of the three methods on perturbations in the first frame -- most challenging as it requires propagating through all timesteps. Here, \tool(LP), \tool(OPT), and POPQORN take 65,348, and 2,160 seconds respectively per example on average. We conclude that both variants of \tool are more precise than POPQORN while being 33.2$\times$ and 6.21$\times$ more scalable for LP and OPT respectively.

\begin{table}[t]
  \centering
  \caption{
    Certification of several LSTM models using \tool with $\epsilon=0.01$. $F$, $H$, and $L$ denote the number of frames,  LSTM hidden units and layers respectively.
  }
  \begin{tabular}{lllrrr}
\toprule
$F$ & $H$ & $L$ & Accuracy (\%) & \shortstack{Certified (\%) \\ by OPT / LP} & \shortstack{Running time (s) \\ by OPT} \\ 
\toprule
\textbf{4} & \textbf{32} & \textbf{1} & 96.1 & 91 / 89 & 14.5 \\ 
\midrule
4 & 32 & \textbf{2} & 96.7 & 92 / 73 & 29.1 \\
4 & 32 & \textbf{3} & 95.8 & 95 / 65 & 43.1 \\ 
\midrule
4 & \textbf{64} & 1 & 97.3 & 93 / 92 & 27.0 \\
4 & \textbf{128} & 1 & 97.1 & 95 / 95 & 52.4 \\ 
\midrule
\textbf{7} & 32 & 1 & 96.5 & 63 / 56 & 32.1 \\ 
\bottomrule
\\
\end{tabular}

  \label{fig:scalability}
  \vspace{-0.5cm}
\end{table}

\textbf{Effect of model size}
We evaluate the scalability of \tool by certifying several recurrent architectures, with varying number of frames $F$, hidden units $H$ and LSTM layers $L$. For each network, we certify the first 100 correctly classified images using the same perturbation $\epsilon=0.01$ for all frames, with 3 repetitions. While in the previous experiment we certified each frame separately to closely follow the setup from \cite{ko2019popqorn}, it is more natural to assume the adversary is able to perturb the entire input.
The results are shown in~\cref{fig:scalability}. We observe that the precision of \tool is affected mostly by the number of frames, as the precision loss accumulates along the frames. Naturally, the running time increases with the number of neurons and frames, as \tool is optimizing the bounds for each $\sigma(x)\tanh(y)$ operation.
However, we also observe a counter-intuitive phenomenon that \tool(OPT) performs better with multi-layer models than with the single-layer model. The precision from \tool(LP) drops with the number of LSTM layers unlike those from \tool(OPT). We hypothesize that an increased number of trainable parameters enhances the flexibility of the bounds for the optimization, allowing us to find more combinations of the bounds that certify the input. \tool(LP) has non-flexible bounds, so the error propagates.

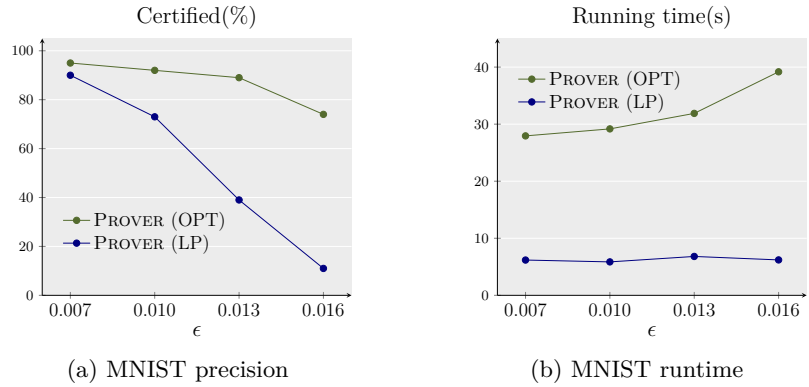
\begin{figure}[t]
  \centering
  \hspace*{0cm}
  \begin{subfigure}{0.4\textwidth}
    \centering
    \scalebox{0.6}{\pgfplotsset{scaled x ticks=false}
\begin{tikzpicture}
  \begin{axis}[
    grid style={draw=white},
    ymajorgrids=true,
    xmin=0.006, xmax=0.017,
    ymin=0,    ymax=105,
    minor tick num=0,
    axis x line=bottom,
    axis y line=left,
    legend style={fill=none, draw=none, at={(0.05, 0.35)}, anchor=north west},
    legend style={fill=none, draw=none},
    legend cell align={left},
    xlabel={\Large $\epsilon$},
    ylabel={\Large Certified(\%)},
    xticklabels={\large0.007, \large0.010, \large0.013, \large0.016, \large0.019},
    xtick=data,
    every axis y label/.style={
      at={(0.5, 1.0)}, above=1mm
    },
    axis background/.style={fill=gray!15}
    ]

    \addplot+ [draw=OliveGreen, mark=*, mark options={scale=1, fill=OliveGreen}]
    table [x=x, y=precision, col sep=comma]{
      x,  precision
      0.007,  95.0
      0.010,  92.0
      0.013,  89.0
      0.016,  74.0
    };

    \addplot+ [draw=NavyBlue, mark=*, mark options={scale=1, fill=NavyBlue}]
    table [x=x, y=precision, col sep=comma]{
      x,  precision
      0.007,  90.0
      0.010,  73.0
      0.013,  39.0
      0.016,  11.0
    };

    \legend{
      \large \tool(OPT),
      \large \tool(LP)};

  \end{axis}
\end{tikzpicture}}
    \caption{MNIST precision}
    \label{fig:mnist_precision}
  \end{subfigure}
  \hspace*{\fill}
  \begin{subfigure}{0.4\textwidth}
    \centering
    \scalebox{0.6}{\pgfplotsset{scaled x ticks=false}
\begin{tikzpicture}
  \begin{axis}[
    grid style={draw=white},
    ymajorgrids=true,
    xmin=0.006, xmax=0.017,
    ymin=0,    ymax=45,
    minor tick num=0,
    axis x line=bottom,
    axis y line=left,
    legend style={fill=none, draw=none, at={(0.05, 0.9)}, anchor=north west},
    legend style={fill=none, draw=none},
    legend cell align={left},
    xlabel={\Large $\epsilon$},
    ylabel={\Large Running time(s)},
    xticklabels={\large0.007, \large0.010, \large0.013, \large0.016, \large0.019},
    xtick=data,
    every axis y label/.style={
      at={(0.5, 1.0)}, above=1mm
    },
    axis background/.style={fill=gray!15}
    ]

    \addplot+ [draw=OliveGreen, mark=*, mark options={scale=1, fill=OliveGreen}]
    table [x=x, y=runtime, col sep=comma]{
      x,  runtime
      0.007,  27.943
      0.010,  29.169
      0.013,  31.885
      0.016,  39.173
    };

    \addplot+ [draw=NavyBlue, mark=*, mark options={scale=1, fill=NavyBlue}]
    table [x=x, y=runtime, col sep=comma]{
      x,  runtime
      0.007,  6.1823
      0.010,  5.8595
      0.013,  6.8133
      0.016,  6.2057
    };
    \legend{
      \large \tool(OPT),
      \large \tool(LP)};

  \end{axis}
\end{tikzpicture}}
    \caption{MNIST runtime}
    \label{fig:mnist_runtime}
  \end{subfigure}
  \hspace*{\fill}
  \caption{Results on MNIST with different epsilons and $F=4,H=32,L=2$.}
  \label{fig:mnist_plot}
  \vspace{-0.6cm}
\end{figure}

\textbf{Effect of perturbation budget}
We certify the robustness of the MNIST classifier for different $\epsilon$ values. We again evaluated 100 correctly classified samples from the test set. \cref{fig:mnist_plot} shows the experimental results. The OPT version has significantly higher precision than LP: i.e., for $\epsilon=0.013$ in \cref{fig:mnist_precision}, LP proves 39\% while OPT certifies 89\% of samples with a higher runtime in \cref{fig:mnist_runtime}.

\subsection{Motion sensor data classification}

We further demonstrate the scalability of \tool by considering a large classifier containing 4 LSTM layers with 256 hidden units each for the human activity recognition dataset HAPT \cite{reyes2016transition}. Each input in the dataset consists of recorded triaxial linear accelerations and angular velocities, sampled at 50Hz.
%
Here, we restricted HAPT to six activity classes and we trimmed angular velocities to at most 6 seconds after the point of prediction.
%
Each input sequence is sliced into sliding windows of 0.5 seconds, which are then passed as an input to the classifier. The trained classifier achieves 88\% test accuracy. Identical to the other experiments, we run \tool on the first 100 correctly classified inputs.

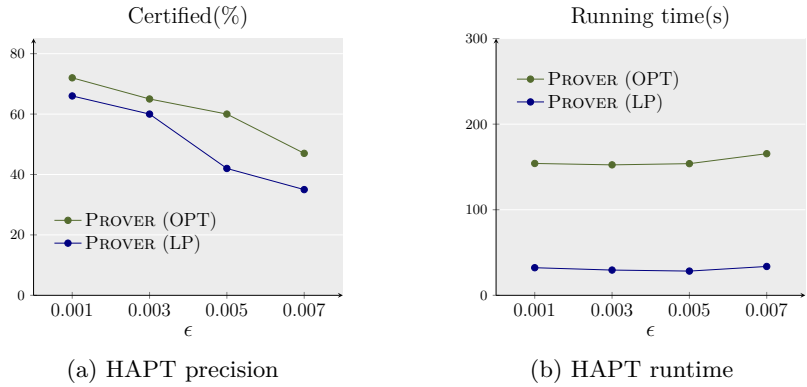
\begin{figure}[t]
  \centering
  \hspace*{0cm}
  \begin{subfigure}{0.4\textwidth}
    \centering
    \scalebox{0.6}{\pgfplotsset{scaled x ticks=false}
\begin{tikzpicture}
  \begin{axis}[
    grid style={draw=white},
    ymajorgrids=true,
    xmin=0.000, xmax=0.008,
    ymin=0,    ymax=85,
    minor tick num=0,
    axis x line=bottom,
    axis y line=left,
    legend style={fill=none, draw=none, at={(0.05, 0.35)}, anchor=north west},
    legend style={fill=none, draw=none},
    legend cell align={left},
    xlabel={\Large $\epsilon$},
    ylabel={\Large Certified(\%)},
    xticklabels={\large0.001, \large0.003, \large0.005, \large0.007},
    xtick=data,
    every axis y label/.style={
      at={(0.5, 1.0)}, above=1mm
    },
    axis background/.style={fill=gray!15}
    ]

    \addplot+ [draw=OliveGreen, mark=*, mark options={scale=1, fill=OliveGreen}]
    table [x=x, y=precision, col sep=comma]{
      x,  precision
      0.001,  72
      0.003,  65
      0.005,  60
      0.007,  47
    };

    \addplot+ [draw=NavyBlue, mark=*, mark options={scale=1, fill=NavyBlue}]
    table [x=x, y=precision, col sep=comma]{
      x,  precision
      0.001,  66
      0.003,  60
      0.005,  42
      0.007,  35
    };

    \legend{
      \large \tool(OPT),
      \large \tool(LP)};

  \end{axis}
\end{tikzpicture}}
    \caption{HAPT precision}
    \label{fig:hapt_precision}
  \end{subfigure}
  \hspace*{\fill}
  \begin{subfigure}{0.4\textwidth}
    \centering
    \scalebox{0.6}{\pgfplotsset{scaled x ticks=false}
\begin{tikzpicture}
  \begin{axis}[
    grid style={draw=white},
    ymajorgrids=true,
    xmin=0.000, xmax=0.008,
    ymin=0,    ymax=300,
    minor tick num=0,
    axis x line=bottom,
    axis y line=left,
    legend style={fill=none, draw=none, at={(0.05, 0.9)}, anchor=north west},
    legend style={fill=none, draw=none},
    legend cell align={left},
    xlabel={\Large $\epsilon$},
    ylabel={\Large Running time(s)},
    xticklabels={\large0.001, \large0.003, \large0.005, \large0.007},
    xtick=data,
    every axis y label/.style={
      at={(0.5, 1.0)}, above=1mm
    },
    axis background/.style={fill=gray!15}
    ]

    \addplot+ [draw=OliveGreen, mark=*, mark options={scale=1, fill=OliveGreen}]
    table [x=x, y=runtime, col sep=comma]{
      x,  runtime
      0.001,  154.073
      0.003,  152.437
      0.005,  153.847
      0.007,  165.483
    };

    \addplot+ [draw=NavyBlue, mark=*, mark options={scale=1, fill=NavyBlue}]
    table [x=x, y=runtime, col sep=comma]{
      x,  runtime
      0.001,  32.256
      0.003,  29.518
      0.005,  28.323
      0.007,  33.723
    };
    \legend{
      \large \tool(OPT),
      \large \tool(LP)};

  \end{axis}
\end{tikzpicture}}
    \caption{HAPT runtime}
    \label{fig:hapt_runtime}
  \end{subfigure}
  \hspace*{\fill}
  \caption{Results on HAPT with different epsilons and $H=256,L=4$.}
  \label{fig:hapt_plot}
  \vspace{-0.6cm}
\end{figure}

Results, shown in \Cref{fig:hapt_plot}, indicate that \tool(OPT) verifies more inputs than \tool(LP), for all perturbation budgets.
Although the number of parameters has increased, \cref{fig:hapt_runtime} shows smaller running times compared to \cref{fig:fsdd_runtime} and \cref{fig:gsc_runtime}. This is because of the smaller number of classes in HAPT, as the verification needs to perform the backsubstitution for each incorrect class. This result shows that \tool (i) is applicable to LSTM classifiers in various domains, and (ii) scales to the large models.

\section{Conclusion}

We introduced a novel approach for certifying RNNs based on a combination of linear programming and abstraction refinement. The key idea is to compute a polyhedral abstraction of the non-linear operations found in the recurrent cells and to dynamically adjust this abstraction according to each input example being certified. Our experimental results show that \tool is more precise and scalable than prior work. These advances enable \tool to certify, for the first time, the robustness of LSTM-based speech classifiers.


\bibliographystyle{splncs04}
\bibliography{bibliography.bib}


\end{document}